\documentclass[journal]{IEEEtran}
\ifCLASSINFOpdf
\else
\fi

\usepackage{mathtools}
\usepackage{graphicx}
\usepackage{url}
\usepackage{hyperref}
\usepackage{booktabs}
\usepackage{breakurl}
\usepackage{multirow}
\usepackage{array}
\usepackage{tabu}
\usepackage[dvipsnames]{xcolor}
\usepackage{amsmath}
\usepackage{amssymb}
\usepackage{comment}
\usepackage{algorithm}
\usepackage[noend]{algpseudocode}
\usepackage{tabularx}

\usepackage{float}
\usepackage{subfig}

\newcolumntype{C}{>{\centering\arraybackslash}p}

\usepackage[flushleft]{threeparttable}
\usepackage{tablefootnote}

\makeatletter
\def\algbackskip{\hskip-\ALG@thistlm}
\makeatother

\usepackage[hang,flushmargin]{footmisc}

\renewcommand{\arraystretch}{1.3}

\makeatletter
\def\footnoterule{\relax%
 \kern-5pt
 \hbox to \columnwidth{\hfill\vrule width \columnwidth height 0.6pt\hfill}
 \kern4.6pt}
\makeatother

\begin{document}
%\title{Learning the Unknowns:\\Towards Fingerprint Spoof Generalization}
\title{White-Box Evaluation of Fingerprint Recognition Systems}

\author{Steven A. Grosz*,~Joshua J.~Engelsma, and Anil K. Jain,~\IEEEmembership{Life~Fellow,~IEEE}%
        
\thanks{S. A. Grosz, J. Engelsma and A. K. Jain are with the Department of Computer Science and Engineering, Michigan State University, East Lansing, MI, 48824. E-mail: \{groszste, engelsm7, jain\}@cse.msu.edu}%
% \thanks{N. G. Paulter is with the National Institute of Standards and Technology, Gaithersburg, Maryland, 20899. E-mail: paulter@nist.gov}%
\thanks{*Corresponding Author}
\thanks{A preliminary version of this paper is to appear in the IEEE International Joint Conference on Biometrics (IJCB), Houston, TX, Sept 28-Oct 1, 2020.~\cite{grosz2019whitebox}.}
}



\markboth{IEEE TRANSACTIONS ON INFORMATION FORENSICS AND SECURITY}%
{Shell \MakeLowercase{\textit{et al.}}: Bare Demo of IEEEtran.cls for IEEE Journals}



\maketitle
\begin{abstract}
    Typical evaluations of fingerprint recognition systems consist of end-to-end black-box evaluations, which assess performance in terms of overall identification or authentication accuracy. However, these black-box tests of system performance do not reveal insights into the performance of the individual modules, including image acquisition, feature extraction, and matching. On the other hand, white-box evaluations, the topic of this paper, measure the individual performance of each constituent module in isolation. While a few studies have conducted white-box evaluations of the fingerprint reader, feature extractor, and matching components, no existing study has provided a full system, white-box analysis of the uncertainty introduced at each stage of a fingerprint recognition system. In this work, we extend previous white-box evaluations of fingerprint recognition system components and provide a unified, in-depth analysis of fingerprint recognition system performance based on the aggregated white-box evaluation results. In particular, we analyze the uncertainty introduced at each stage of the fingerprint recognition system due to adverse capture conditions (i.e., varying illumination, moisture, and pressure) at the time of acquisition. Our experiments show that a system that performs better overall, in terms of black-box recognition performance, does not necessarily perform best at each module in the fingerprint recognition system pipeline, which can only be seen with white-box analysis of each sub-module. Findings such as these enable researchers to better focus their efforts in improving fingerprint recognition systems.
\end{abstract}


\IEEEpeerreviewmaketitle

\section{Introduction}
\label{sec:introduction}

\begin{figure}
	\centering
	\includegraphics[width=0.48\textwidth]{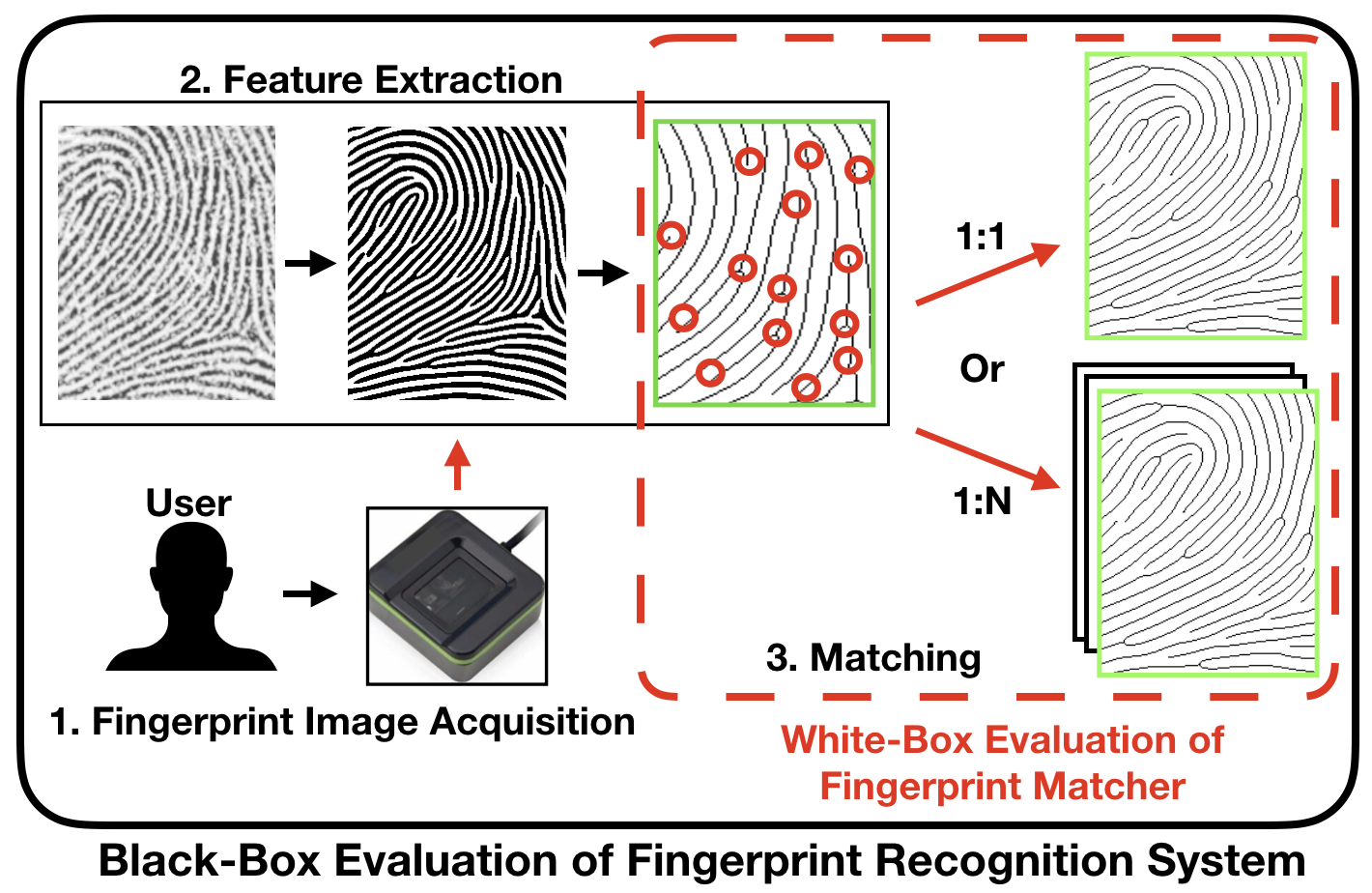}%
	\caption{Overview of the various modules of an automated fingerprint recognition system. Typical performance evaluations are conducted in an end-to-end, black-box manner. In contrast, a white-box evaluations, such as for the matcher component highlighted in red, assess performance at the sub-module level.}
	\label{fig:system_diagram}
\end{figure}

\IEEEPARstart{M}{ost} techniques for evaluating automated fingerprint identification systems (AFIS) consist of a black-box evaluation of authentication or search accuracy on a given dataset.\footnote{Black-box testing focuses on testing the end-to-end system using inputs and outputs (e.g., fingerprint image and score, respectively)~\cite{testing}. In contrast, white-box testing evaluates the internal sub-components of a system.} For example, the National Institute of Standards and Technology (NIST) conducts fingerprint vendor technology evaluations (FpVTE)~\cite{nist} and the University of Bologna conducts fingerprint verification competitions (FVC) (\cite{maio2002fvc2002},~\cite{maio2004fvc2004},~\cite{dorizzi2009fingerprint}) to evaluate fingerprint recognition systems, as measured in terms of computational requirements and recognition accuracy on benchmark datasets. Black-box evaluations are valuable in that they allow for overall comparisons between recognition systems in terms of operational performance. However, black-box approaches are limited in that they lack granularity into the performance of the individual sub-modules of the system (image acquisition, feature extraction, and matching) shown in Figure~\ref{fig:system_diagram}. 

Several white-box studies, aimed at evaluating the various sub-components of an AFIS (fingerprint reader, minutiae extractors, and minutiae matchers), have been proposed to address this limitation inherent to black-box evaluations~\cite{3Dfingers, whole_hand, arora2017gold, engelsma2018universal, chugh2017benchmarking, grosz2019whitebox}. While these studies provide a good starting point for white-box evaluations of each constituent AFIS module, we posit that (i) each of them leaves room for more rigorous testing and uncertainty analysis and also (ii) it is more beneficial to perform all of these evaluations together in a unified framework in order to gain additional insights. Therefore, in this paper, we propose extensive white-box evaluations and an uncertainty analysis of each of the major components (fingerprint reader, minutiae extractors, and minutiae matchers) of two state-of-the-art AFIS. Then, we examine these individual white-box evaluation results in light of an end-to-end black-box evaluation of the two AFIS. The contributions made to the white-box evaluations of each component of AFISs, along with our unified end-to-end evaluation are described in the following subsections.

\subsection{\textbf{Fingerprint Reader Evaluation}}

Perhaps the most well known evaluations for fingerprint readers are enumerated in the certification standards PIV-071006~\cite{nill2006test} and Appendix F~\cite{nill2005test}. A limitation with these evaluations is that they only utilize calibration patterns for evaluation. However, these patterns are significantly different from the fingerprints the readers will be sensing in an operational scenario. This limitation prompted studies in~\cite{3Dfingers, whole_hand, arora2017gold, engelsma2018universal} to develop realistic, 3D fingerprint targets to evaluate fingerprint readers in a more operational setting. These studies did utilize the 3D targets for some white-box testing, but were primarily focused on evaluations related to end-to-end matching performance. This is also the case with~\cite{cappelli2008operational}, where the authors investigate the effect of relaxing each metric outlined in the PIV-071006 in terms of black-box matching accuracy on a database of real fingerprints.

Other studies have performed a more isolated evaluation of the fingerprint reader module (white-box evaluation) through a variety of fingerprint image quality-metrics (\cite{lim2002fingerprint,ridge_count,chen2005fingerprint_quality,tabassi2004fingerprint, tabassi2005novel}). For example, in~\cite{alonso2008performance}, the performance of both an optical and capacitive reader are evaluated using a variety of fingerprint quality-metrics in the presence of varying amounts of finger pressure and moisture. Similarly, four types of fingerprint sensing technologies, including optical, semiconductor, thermal, and tactile, are evaluated (again using fingerprint quality-metrics) by Kang \textit{et al.} in~\cite{kang2003study}.
    
In this work, we build upon the white-box fingerprint reader evaluations of~\cite{alonso2007comparative, kang2003study} by including an additional sensing technology (ultrasound), an additional capture condition (illumination), and a statistical uncertainty analysis/evaluation to better determine the sensitivity of the fingerprint readers to the varying conditions. In total, we evaluate fingerprint readers equipped with three sensing technologies (capacitive, optical, and ultrasound) under three varying capture conditions (humidity, pressure, and lighting). Our white-box fingerprint reader evaluation provides a much needed, comprehensive update on the state of fingerprint sensing technologies, which have likely progressed since the study of Alonso-Fernandez \textit{et al.}~\cite{alonso2008performance} over a decade ago.

\subsection{\textbf{Minutiae Extractor Evaluation}}

Chugh \textit{et al.} conduct a white-box evaluation of minutiae extractors in~\cite{chugh2017benchmarking}. In particular, they measure the detection and localization performance of four minutiae-based feature extractors in the presence of random perturbations to the input fingerprint images. A major limitation of this study is that the types of perturbations, namely specular noise and motion blur, are not an adequate model of perturbations that exist in the real world (e.g., dry/wet fingers and distortion due to varying pressure). 

Therefore, in this work, we devise a more robust white-box evaluation of the minutiae extraction module via more realistic perturbation techniques. In particular, rather than randomly adding noise or motion blur to fingerprint images prior to minutiae extraction, we instead apply techniques from neural style transfer to add a ``moisture style" or a ``pressure style" as a perturbation to an input fingerprint prior to performing minutiae extraction. In this manner, we are able to much better independently and uniquely evaluate the performance of minutiae extractors in the presence of realistic perturbations. As with the fingerprint reader module, our evaluation of the minutiae extraction module also uses a statistical uncertainty analysis to determine the sensitivity of the minutiae extractors to these real world perturbations.

\subsection{\textbf{Minutiae Matcher Evaluation}}

Grosz \textit{et al.} performed the first white-box analysis of minutiae-based matchers in~\cite{grosz2019whitebox}. In particular, the authors in~\cite{grosz2019whitebox} evaluated the sensitivity of one open-source and two commercial-off-the-shelf (COTS) minutiae-matchers to various random perturbations and statistically modeled (using fingerprint training data) non-linear distortions of the input minutiae feature sets. 

We build upon the work of~\cite{grosz2019whitebox} by adding additional perturbations to the minutiae set that are more cognizant of the types of perturbations that can be realistically encountered in an operational setting. For example, rather than removing minutiae randomly throughout the fingerprint image as was done in~\cite{grosz2019whitebox}, we simulate perturbations to the minutiae set that stem from wet impressions by removing spatially contiguous blocks of minutiae points throughout the minutiae set (since wet prints often result in a collapsed ridge structure in certain blocks of the image and consequently a missing block of minutiae points). We also investigate global rotations of the minutiae sets to simulate varying angles at which a finger is placed during multiple acquisitions. With these additional, more realistic minutiae perturbations, we conduct an uncertainty analysis on two state-of-the-art minutiae matchers to determine their sensitivity to these new perturbation techniques. 

\subsection{\textbf{End-to-End Evaluation}}

After conducting rigorous white-box evaluations and uncertainty analysis on each of the AFIS modules in isolation, we conclude by conducting a full end-to-end evaluation of each AFIS (two state-of-the-art COTS AFIS). This unified combination of both rigorous white-box and black-box evaluations serve as a more comprehensive and complete AFIS evaluation than existing methods. By combining a full black-box evaluation together with the individual white-box evaluations, we show that even though one AFIS may perform better overall than another in a black-box evaluation, it might not be the case that each individual sub-module of that system is best. This serves as motivation for the various standard evaluations such as the NIST FpVTE to adopt new white-box evaluation techniques to more thoroughly investigate each vendor's AFIS. 

Our combination of white-box and black-box evaluations also enables researchers and engineers to select the best combination of AFIS components suitable for their application domain. For example, a buyer interested in an AFIS which needs to operate in an outdoor environment with high illumination and high humidity (e.g., in a developing country) may prefer to purchase a fingerprint reader from one company and a feature extractor and matcher from another company, such that both modules have little sensitivity to illumination and moisture, while also making sure that the components can be integrated together to obtain high end-to-end (black-box) performance.

In summary, the contributions of this research are:

\begin{itemize}
    \item More complete and rigorous white-box evaluation approaches than have previously been conducted for individual AFIS modules including fingerprint readers, feature extractors (minutiae extractors), and matchers (minutiae matchers). We accomplish this via additional perturbation techniques that more closely approximate real world perturbations, an in-house fingerprint dataset collected in the presence of various perturbations (moisture, pressure, and illumination), and an uncertainty analysis evaluation protocol. 
    
    \item A white-box evaluation on three commercial, FBI certified fingerprint readers, each employing different sensing technology (optical, capacitive, and ultrasound) on fingerprint impressions captured (at MSU) under varying illumination, finger pressure on the platen, and finger dryness.
    
    \item A white-box evaluation of two state-of-the-art, COTS minutiae extractors. We show how neural style transfer can be used to add highly realistic perturbations to fingerprint images (e.g., moisture perturbations) to study the sensitivity of minutiae extraction performance.
    
    \item A white-box evaluation of two state-of-the-art COTS minutiae matchers. We extend our prior white-box minutiae-matcher evaluation~\cite{grosz2019whitebox} by assessing robustness to two additional (more realistic) perturbations not previously studied: global rotation and occlusion of minutiae features sets.
    
    \item A unified framework which enables module level evaluations and also interpretation of these modular evaluation in light of end-to-end black-box performance. The code for these experiments is available at~\url{https://github.com/groszste/AFIS-WhiteBoxEvaluation}
\end{itemize}

The organization of the rest of this papers is as follows. In section 2, we present the white-box analysis of the fingerprint reader module of an AFIS to various capture perturbations; in section 3, we do the same for the feature extractor module; and this is followed by an analysis of the matcher component in section 4. Within each of these sections, we detail the evaluation procedure, datasets used, experimental protocols, and experimental results. Finally, section 5 complements the white-box evaluations of the individual modules with an end-to-end black-box evaluation of a given fingerprint recognition system. Section 6 then concludes the paper with a summary of the results and a discussion on future directions related to this work.

\section{Datasets}

In our white-box and black-box evaluations, we use a number of different fingerprint datasets. In addition to a number of publicly available datasets, we collected our own dataset which is well suited for our white-box evaluations. We call this dataset the Varying Capture Conditions (VCC) dataset.

\subsection{VCC Dataset}

The VCC dataset is comprised of $3421$ fingerprints acquired under different pressure, moisture, and illumination on three different fingerprint readers: one frustrated total internal refection (FTIR) optical-based reader, one capacitive-based reader, and one ultrasound-based reader. The sub-categories of adverse capture conditions included dry finger, unaltered finger moisture, and wet finger (for the moisture condition), bright lighting, normal lighting, and dark lighting (for the illumination condition), and high pressure, medium pressure, and low pressure (for the pressure condition). Table~\ref{tab:data_ranges} gives quantitative measurement ranges for each capture condition.

% \begin{figure}[t]
% 	\centering
% 	\subfloat[]{\includegraphics[width=0.225\textwidth]{images/sensors/morpho.png}%
% 	\label{1}}
% 	\hfil
% 	\subfloat[]{\includegraphics[width=0.225\textwidth]{images/sensors/qualcomm.png}%
% 	\label{2}}
% 	\hfil
% 	\subfloat[]{\includegraphics[width=0.225\textwidth]{images/sensors/upek.jpg}%
% 	\label{3}}
% 	\caption{Images of the three fingerprint readers used in this study. (a) MorphoTop optical-FTIR reader, (b) Qualcomm SnapDragon ultrasound reader, and (c) Digital Persona EikonTouch 710 capacitive reader.}
% 	\label{fig:sensors}
% \end{figure}
        
To control the ambient illumination for these experiments, we used a digital light meter and an external light source placed directly above each reader's imaging surface\footnote{We used a Halogen 80-Watt 1600-Lumen PAR 38 Floodlight bulb.}. For bright illumination, the light source is placed at a vertical distance such that the illumination recorded at the imaging surface is $50\,000$ Lux. For dark lighting, a covering is placed over the fingerprint readers to shade the imaging surface such that $10$ Lux is recorded. For the normal capture condition, the ambient office environment lighting was recorded as 250 Lux.
        
To control the finger moisture content, a skin moisture measurement device was used to record the percentage of moisture on the skin surface. The categories of moisture recorded were the subject's natural skin moisture (normal condition), the moisture after wiping the fingertips with a dry paper towel (dry condition), and after applying a small amount (roughly $0.25$ g) of moisturizing lotion to the fingertips (\textit{i.e.}, wet condition). Since each subject has a different natural moisture content, the average and standard deviation of moisture content for each category are provided in Table~\ref{tab:data_ranges}.
        
Finally, we use a tactile grip force and pressure sensing system to measure the amount of pressure applied under the three settings of normal pressure, low pressure, and high pressure. For the normal pressure scenario, participants were asked to present their fingerprints to the readers without intentionally controlling the amount of pressure. For low pressure, subjects were asked to present each fingerprint by simply placing their fingers on the imaging surface, doing their best not to apply any downward force. Lastly, subjects were asked to present their fingers to the imaging surface while applying their maximum downward force for high pressure impressions. Measurement ranges for these three categories are presented in Table~\ref{tab:data_ranges}.

\begin{table}[t]
	\centering
	\caption{Measurement ranges for each condition of the Varying Capture Conditions (VCC) Dataset}
	\label{tab:data_ranges}
	\small 
	{
	\begin{tabular}{p{0.26\linewidth-2\tabcolsep}C{0.2\linewidth-2\tabcolsep}C{0.27\linewidth-2\tabcolsep}C{0.27\linewidth-2\tabcolsep}}
		\toprule
      		& Illumination \newline (Lux) & Pressure \newline (kPA) & Skin Moisture \newline (\%) \\
		\midrule
	    Normal & $250$ & $38.8 \pm 12.14$ & $76.23 \pm 6.68$ \\
		\midrule
	    Dry finger & $250$ & $38.8 \pm 12.14$ & $68.6 \pm 3.87$ \\
		\midrule
	    Wet finger & $250$ & $38.8 \pm 12.14$ & $90.83 \pm 9.49$ \\
		\midrule
	    Low pressure & $250$ & $17.78 \pm 6.08$ & $76.23 \pm 6.68$\\
		\midrule
	    High pressure & $250$ & $174.38 \pm 49.18$ & $76.23 \pm 6.68$\\
		\midrule
	    Bright lighting & $50\,000$ & $38.8 \pm 12.14$ & $76.23 \pm 6.68$\\
		\midrule
	    Dark lighting & $10$ & $38.8 \pm 12.14$ & $76.23 \pm 6.68$\\
		\bottomrule
	\end{tabular}
	}
\end{table}
        
\begin{figure*}
\centering
\begin{tabular}{p{0.015\textwidth}*{7}{>{\centering\arraybackslash}m{0.1\textwidth}}}
\toprule
 & Normal & Dry & Wet & Low \newline Pressure & High \newline Pressure & Bright \newline Lighting & Dark \newline Lighting \\
 
\midrule

\rotatebox[origin=c]{90}{Optical} & 
\subfloat[]{\includegraphics[width=0.1\textwidth]{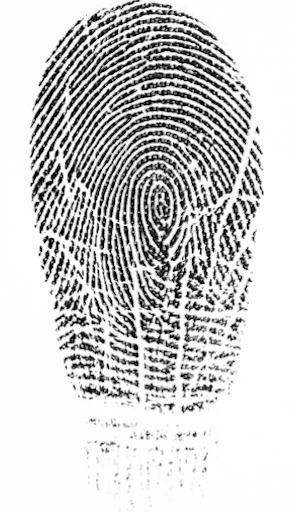}%
\label{morpho normal}} &
\subfloat[]{\includegraphics[width=0.1\textwidth]{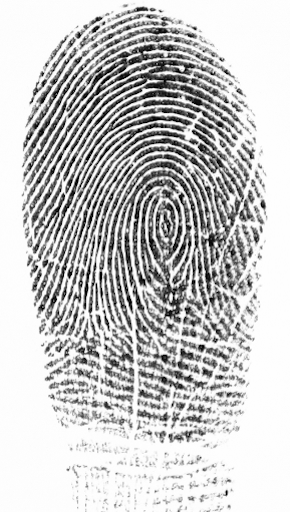}%
\label{morpho dry}} & 
\subfloat[]{\includegraphics[width=0.1\textwidth]{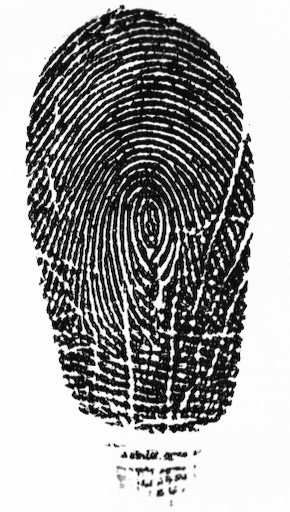}%
\label{morpho wet}} & 
\subfloat[]{\includegraphics[width=0.1\textwidth]{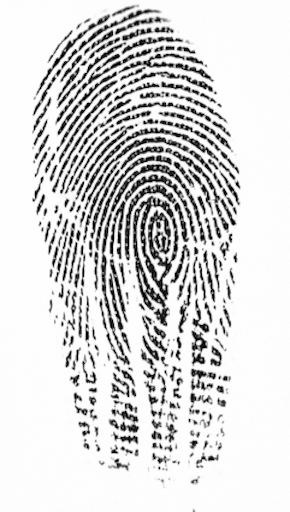}%
\label{morpho low}} & 
\subfloat[]{\includegraphics[width=0.1\textwidth]{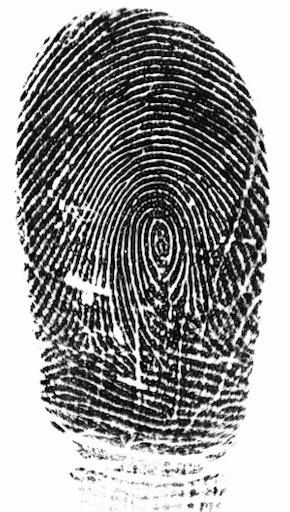}%
\label{morpho high}} & 
\subfloat[]{\includegraphics[width=0.1\textwidth]{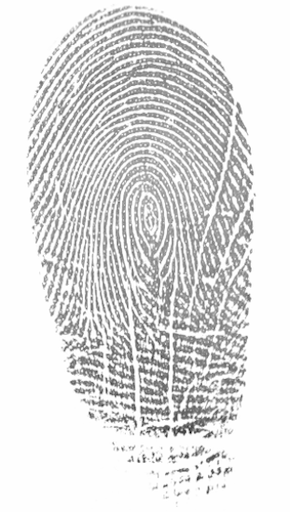}%
\label{morpho bright}} & 
\subfloat[]{\includegraphics[width=0.1\textwidth]{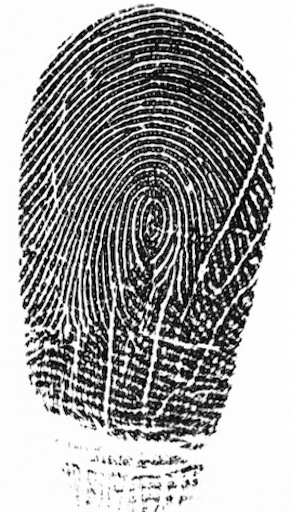}%
\label{morpho dark}} \\

\midrule

\rotatebox[origin=c]{90}{Ultrasound} & 
\subfloat[]{\includegraphics[width=0.1\textwidth]{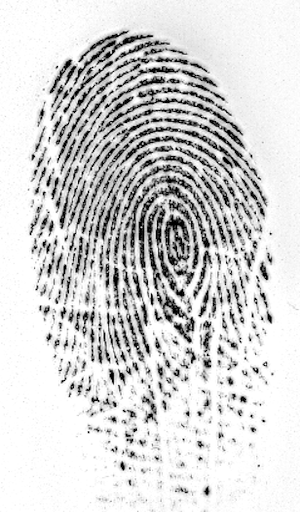}%
\label{qualcomm normal}} &
\subfloat[]{\includegraphics[width=0.1\textwidth]{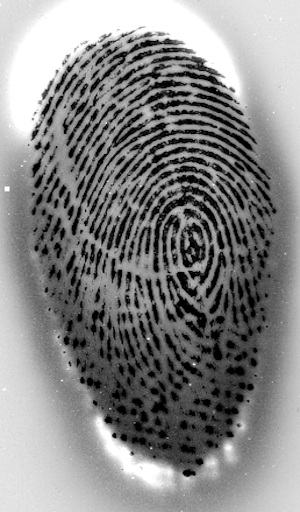}%
\label{qualcomm dry}} & 
\subfloat[]{\includegraphics[width=0.1\textwidth]{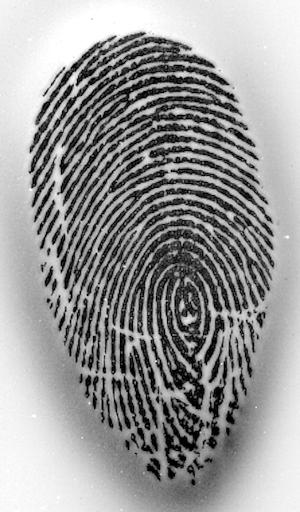}%
\label{qualcomm wet}} & 
\subfloat[]{\includegraphics[width=0.1\textwidth]{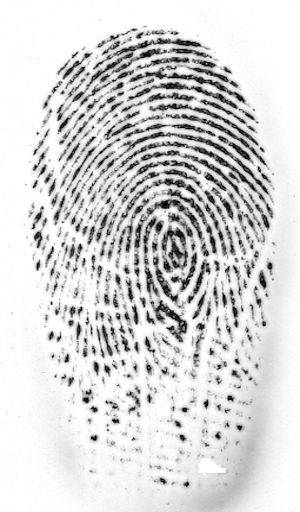}%
\label{qualcomm low}} & 
\subfloat[]{\includegraphics[width=0.1\textwidth]{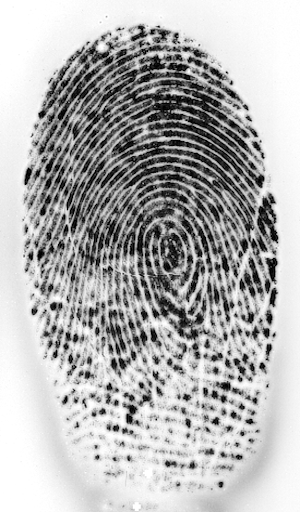}%
\label{qualcomm high}} & 
\subfloat[]{\includegraphics[width=0.1\textwidth]{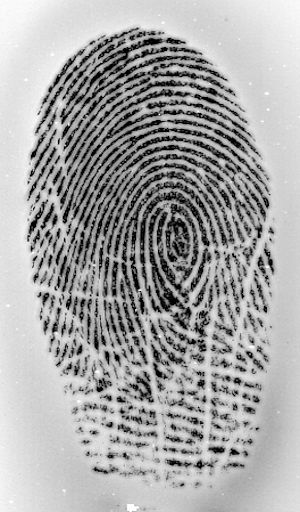}%
\label{qualcomm bright}} & 
\subfloat[]{\includegraphics[width=0.1\textwidth]{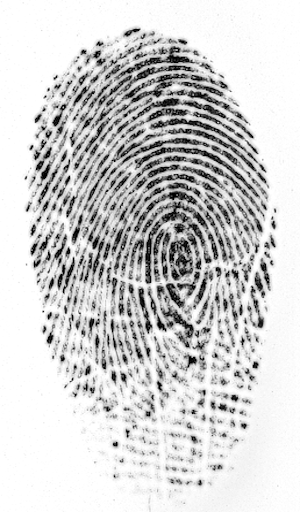}%
\label{qualcomm dark}} \\

\midrule

\rotatebox[origin=c]{90}{Capacitive} & 
\subfloat[]{\includegraphics[width=0.1\textwidth]{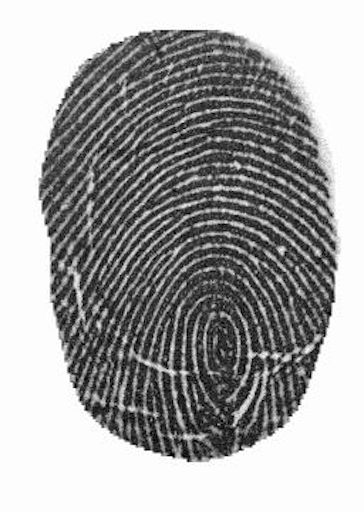}%
\label{upek normal}} &
\subfloat[]{\includegraphics[width=0.1\textwidth]{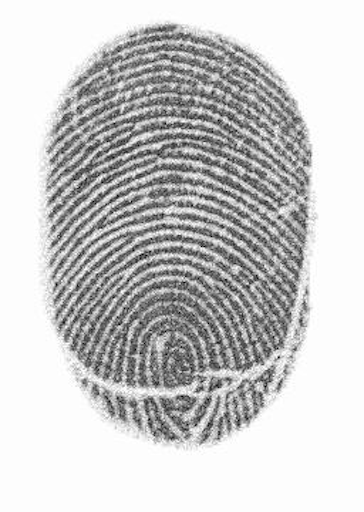}%
\label{upek dry}} & 
\subfloat[]{\includegraphics[width=0.1\textwidth]{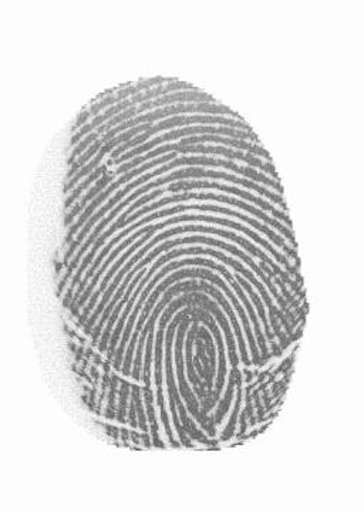}%
\label{upek wet}} & 
\subfloat[]{\includegraphics[width=0.1\textwidth]{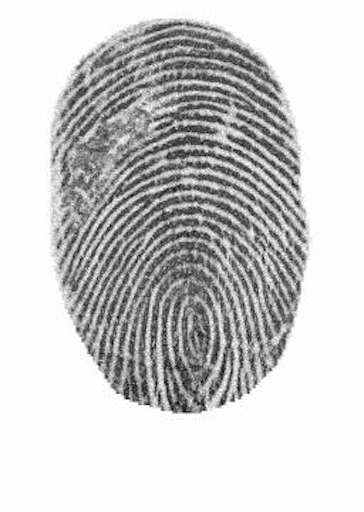}%
\label{upek low}} & 
\subfloat[]{\includegraphics[width=0.1\textwidth]{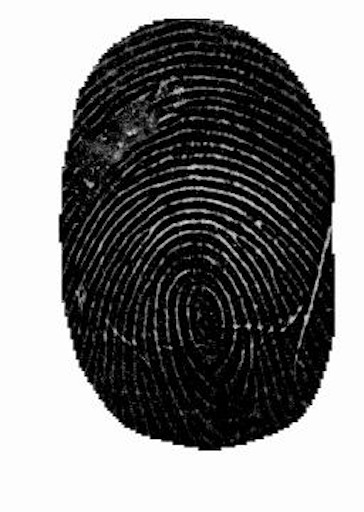}%
\label{upek high}} & 
\subfloat[]{\includegraphics[width=0.1\textwidth]{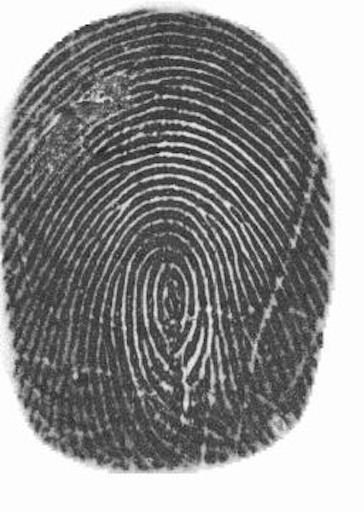}%
\label{upek bright}} & 
\subfloat[]{\includegraphics[width=0.1\textwidth]{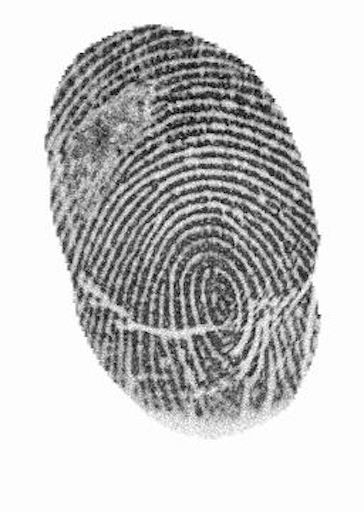}%
\label{upek dark}} \\

\bottomrule
\end{tabular}
\caption{Example fingerprint impressions from each reader captured under $7$ different capture conditions.}
\label{fig:data}
% \vspace{-1em}
\end{figure*}

For each fingerprint, participants were first asked to present each of their ten fingers under ``normal" conditions, i.e., asking the users to press their finger against the imaging surface in a natural way, under ambient office environment lighting, with normal pressure and natural skin humidity. Next, additional impressions of the ten fingers were acquired under the six different conditions with varying sequential order between participants to control for differences in presentations as volunteers became more familiar with placing their fingers on the reader platens. The order of fingerprint readers presented to the volunteers was also varied for the same reason. Due to a high failure to enroll rate and the length of the entire capture session, only eleven participants were asked to image on the capacitive reader. Table~\ref{tab:data_stats} summarizes the statistics of the Varying Capture Conditions (VCC) dataset. Example fingerprints in the presence of each condition are shown in Figure~\ref{fig:data}. The fingerprints were collected from $20$ students, ages 20 to 30 years old with varying ethnic origin (Caucasian, East Asian, and South Asian), from collaborating research labs at MSU (Pattern Recognition and Image Processing Lab, Computer Vision Lab, and Human Analysis Lab).

\begin{table}[t]
	\centering
	\caption{Statistics of the Varying Capture Conditions (VCC) Dataset}
	\label{tab:data_stats}
	\small 
	{
	\begin{tabular}{p{0.12\textwidth}p{0.09\textwidth}p{0.09\textwidth}p{0.09\textwidth}}
		\toprule
%       		& Morpho & Qualcomm & Digital \newline Persona \\
% 		\midrule
		Type & Optical & Ultrasound & Capacitive \\
		\midrule
		Certification & Appendix F & Appendix F & PIV-071006 \\ 
		\midrule
		Image Size \newline (pixels) & $512\times512$ & Varies &  $256\times360$ \\
		\midrule
		Resolution \newline (dpi) & $500$ & $508$ & $508$ \\
		\midrule
		\# subjects & $20$ & $20$ & $11$ \\
		\midrule
		\# normal & $199$ & $190$ & $106$ \\
		\midrule
		\# dry finger & $190$ & $190$ & $106$ \\
		\midrule
		\# wet finger & $199$ & $190$ & $104$ \\
		\midrule
		\# low pressure & $183$ & $190$ & $106$ \\
		\midrule
		\# high pressure & $199$ & $180$ & $106$ \\
		\midrule
		\# bright lighting & $194$ & $190$ & $106$ \\
		\midrule
		\# dark lighting & $197$ & $190$ & $106$ \\
		\bottomrule
		\multicolumn{4}{l}{Note: Variations in \# of impressions are due to failures to enroll.}\\
	\end{tabular}
	}
	\vspace{-1.6em}
\end{table}

\subsection{Public Datasets}

In addition to our internally collected VCC dataset, we also leverage several publicly available datasets to complete our full evaluation protocol.

% CrossMatch\footnote{https://www.hidglobal.com/products/readers/tenprint-readers/guardian-patrol}

(i) We aggregate fingerprint data captured on FTIR optical-based readers from a number of different sources, including FVC 2004 DB1-A~\cite{maio2004fvc2004}, LivDet 2015~\cite{marcialislivdet}, MSU-FPAD~\cite{chugh2018fingerprint}, and various IARPA Governmental Controlled Tests (GCT1, GCT2, and GCT3)\footnote{https://www.iarpa.gov/index.php/research-programs/odin.}. Note that the MSU-FPAD and IARPA GCT data were collected with the intention of training and evaluating fingerprint presentation attack detection algorithms, but we can easily discard the spoof data and utilize the live data for our white-box evaluations. In total, we have $16\,731$ fingerprint images in this aggregated dataset. In our experiments, we use this dataset to train our neural style transfer network (our realistic perturbation technique that we apply prior to minutiae extraction and evaluation). 

(ii) Finally, we utilize the FVC 2002 DB1A~\cite{maio2002fvc2002} dataset which has manually annotated ground truth minutiae locations and orientations (provided by Kayaoglu \textit{et al.}~\cite{kayaoglu2013standard}). This dataset consists of $800$ images from $100$ unique fingers collected on an FTIR optical-based reader. The images in this dataset are used to independently evaluate our minutiae extractors and matchers.  
        
\section{White-Box Evaluation Protocol}

For each of our white-box evaluations (fingerprint reader, minutiae extractor, and matcher), we perform a rigorous uncertainty analysis to determine the sensitivity of each module to our realistic perturbations. The particular uncertainty calculation that we use is the Monte Carlo method~\cite{jcgm} for estimating uncertainties. It demonstrates the sensitivity of a module to perturbations of its inputs. A lower uncertainty score is better as it indicates better robustness to the perturbations. 

The step-by-step procedure used to calculate the uncertainty of a module to the controlled, realistic perturbations (\textit{e.g.}, adverse capture conditions) is as follows:

\begin{enumerate}
    \item Generate $M$ number of $A_{k}$ reference fingerprint impressions, $1 \leq k \leq M$.
    \item Obtain $M$ feature sets, $S_{k}$, from each $A_{k}$.
    \item For each $S_{k}$, synthesize $N$ number of perturbed feature sets, $S'_{k,n}$, $1 \leq n \leq N$.
    \item Generate module specific evaluation scores, $s_{k,n}$, between $S_{k}$ and each $S'_{k,n}$.
    \item Normalize the scores, $s_{k,n}$, to be in the range of $[0,1]$ using min-max normalization, where the min and max are module specific values.
    \item Compute the average, $\mu_{k}$, of the $s_{k,n}$ scores using $\mu_{k} = \frac{1}{N}\sum_{n=1}^{N}(s_{k,n})$
    \item Compute the standard uncertainty, $u_k$, of $A_{k}$ using $u_{k} = \sqrt{\frac{1}{N}\sum_{n=1}^{N}(\mu_{k}-s_{k,n})^{2}}$
    \item Repeat steps $1$ to $7$ for each reference feature set obtaining an $u_k$ for each $S_{k}$.
    \item Compute the total uncertainty, $u_{total}$, using $u_{total} = \sqrt{\frac{1}{M}\sum_{k=1}^{M}u_k^2}$
\end{enumerate}

In the following sections, we apply this uncertainty analysis, along with other module specific evaluations to each module of our two COTS AFIS.

% \begin{table*}[t]
% \renewcommand{\arraystretch}{1.3}
% \caption{Uncertainty Scores for MorphoTop, Qualcomm SnapDragon, and Digital Persona EikonTouch 710 Fingerprint Readers.}
% \label{tab:uncertainty_readers}
% \centering
% \begin{tabular}{p{0.1\linewidth-2\tabcolsep}*{4}{C{0.07\linewidth-2\tabcolsep}C{0.08\linewidth-2\tabcolsep}C{0.075\linewidth-2\tabcolsep}}}
% \toprule
%  & \multicolumn{4}{c}{Morpho} & \multicolumn{4}{c}{Qualcomm} & \multicolumn{4}{c}{Digital Persona}\\
% % \midrule
%  & GOQ & RF & OCL & Veri & GOQ & RF & OCL & Veri & GOQ & RF & OCL & Veri \\
% \cmidrule{1-1} \cmidrule(rl){2-5} \cmidrule(rl){6-9} \cmidrule(rl){10-13}
% % \midrule
% Finger Moisture & 0.0084 & 0.0097 & 0.0052 & 0.0105 & 0.0033 & 0.0049 & 0.0091 & 0.0052 & 0.0091 & 0.0097 & 0.0080 & 0.0077 \\
% \cmidrule{1-1} \cmidrule(rl){2-5} \cmidrule(rl){6-9} \cmidrule(rl){10-13}
% Contact Pressure & 0.0094 & 0.0088 & 0.0138 & 0.0143 & 0.0070 & 0.0093 & 0.0065 & 0.0083 & 0.0208 & 0.0110 & 0.0108 & 0.0093 \\
% \cmidrule{1-1} \cmidrule(rl){2-5} \cmidrule(rl){6-9} \cmidrule(rl){10-13}
% Illumination & 0.0169 & 0.0077 & 0.0263 & 0.0058 & 0.0062 & 0.0110 & 0.0027 & 0.0020 & 0.0137 & 0.0059 & 0.0074 & 0.0069 \\
% % \cmidrule{1-1} \cmidrule(rl){2-5} \cmidrule(rl){6-9} \cmidrule(rl){10-13}
% % Average & 0.0116 & 0.0087 & 0.0151 & 0.0102 & 0.0055 & 0.0084 & 0.0061 & 0.0052 & 0.0145 & 0.0089 & 0.0087 & 0.0080 \\
% \bottomrule
% \end{tabular}
% % \vspace{-1.6em}
% \end{table*}

\begin{table*}[t]
\renewcommand{\arraystretch}{1.3}
\caption{Uncertainty Scores for Three Fingerprint Readers Under Varying Capture Conditions.}
\label{tab:uncertainty_readers}
\centering
\begin{tabular}{p{0.1\linewidth-2\tabcolsep}*{4}{C{0.06\linewidth-2\tabcolsep}C{0.08\linewidth-2\tabcolsep}C{0.08\linewidth-2\tabcolsep}}}
\toprule
 & \multicolumn{3}{c}{GOQ} & \multicolumn{3}{c}{RF} & \multicolumn{3}{c}{OCL} & \multicolumn{3}{c}{COTS}\\
% \midrule
 & Optical & Ultrasound & Capacitive & Optical & Ultrasound & Capacitive & Optical & Ultrasound & Capacitive & Optical & Ultrasound & Capacitive \\
\cmidrule{1-1} \cmidrule(rl){2-4} \cmidrule(rl){5-7} \cmidrule(rl){8-10} \cmidrule(rl){11-13}
% \midrule
Finger Moisture & 0.0084 & \textbf{0.0033} & 0.0091 & 0.0097 & \textbf{0.0049} & 0.0097 & \textbf{0.0052} & 0.0091 & 0.0080 & 0.0105 & \textbf{0.0052} & 0.0077 \\
\cmidrule{1-1} \cmidrule(rl){2-4} \cmidrule(rl){5-7} \cmidrule(rl){8-10} \cmidrule(rl){11-13}
Contact Pressure & 0.0094 & \textbf{0.0070} & 0.0208 & \textbf{0.0088} & 0.0093 & 0.0110 & 0.0138 & \textbf{0.0065} & 0.0108 & 0.0143 & \textbf{0.0083} & 0.0093 \\
\cmidrule{1-1} \cmidrule(rl){2-4} \cmidrule(rl){5-7} \cmidrule(rl){8-10} \cmidrule(rl){11-13}
Illumination & 0.0169 & \textbf{0.0062} & 0.0137 & 0.0077 & 0.0110 & \textbf{0.0059} & 0.0263 & \textbf{0.0027} & 0.0074 & 0.0058 & \textbf{0.0020} & 0.0069 \\
\bottomrule
\end{tabular}
% \vspace{-1.6em}
\end{table*}

\section{Fingerprint Reader Evaluation}

To perform our white-box fingerprint reader evaluation, we utilize our VCC dataset along with a variety of fingerprint quality-scores (scores which evaluate the ridge-valley structure of the fingerprint image). In particular, we use the following quality-metrics\footnote{We used open source implementations for GOQ and Ridge Frequency algorithms~\cite{KovesiMATLABCode}}: Orientation Certainty Level (OCL)~\cite{lim2002fingerprint}, Ridge Frequency Estimation (RF)~\cite{lim2002fingerprint}, Global Orientation Quality (GOQ)~\cite{lim2002fingerprint}, and the proprietary quality estimation algorithm of one popular COTS AFIS. These quality-metrics are computed on the fingerprints captured in the VCC dataset to determine the sensitivity of the three fingerprint readers to moisture, illumination, and pressure.
    	
\begin{figure*}
	\centering
	\subfloat[]{\includegraphics[width=0.33\textwidth]{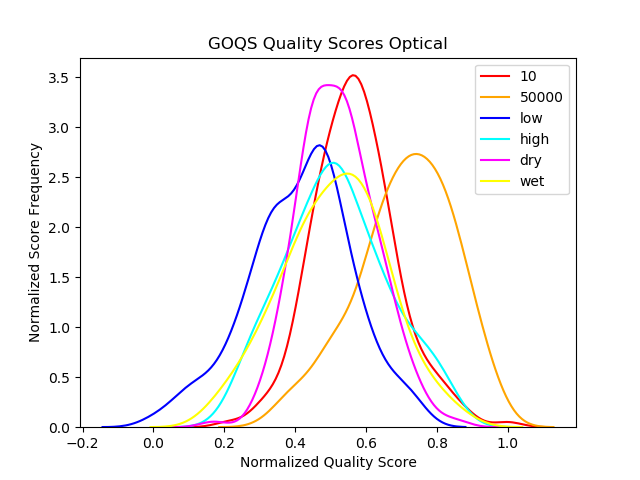}%
	\label{1case}}
	\hfil
	\subfloat[]{\includegraphics[width=0.33\textwidth]{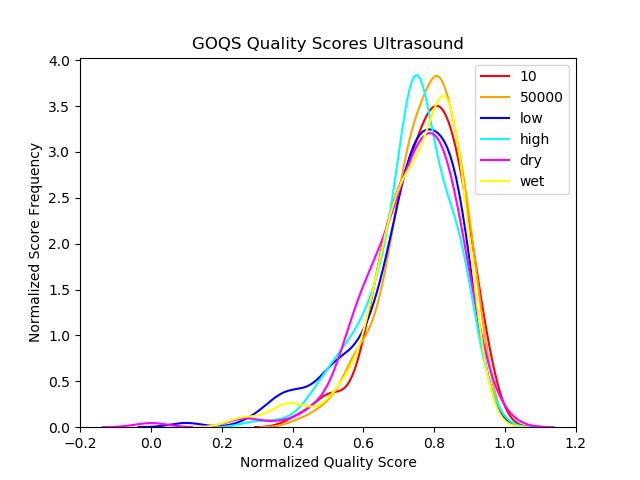}%
	\label{2case}}
	\hfil
	\subfloat[]{\includegraphics[width=0.33\textwidth]{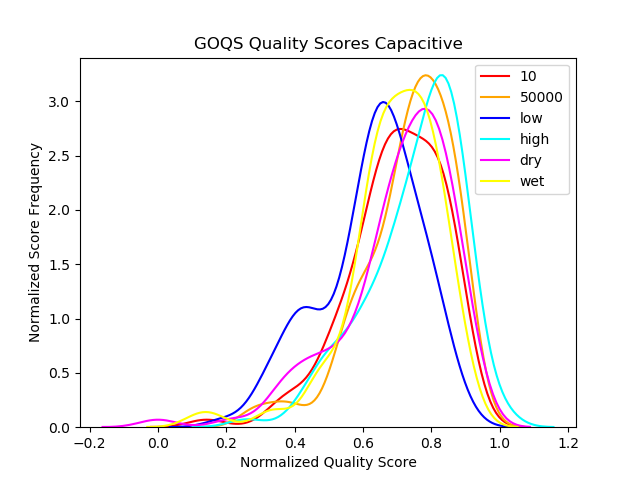}%
	\label{3case}}
	
	\hfil
	\subfloat[]{\includegraphics[width=0.33\textwidth]{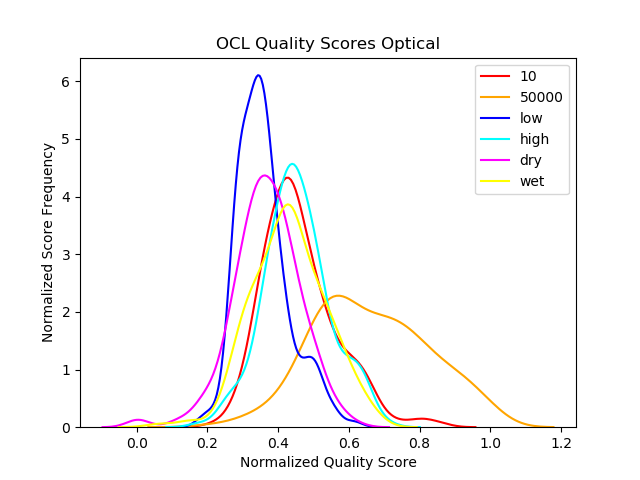}%
	\label{4case}}
	\hfil
	\subfloat[]{\includegraphics[width=0.33\textwidth]{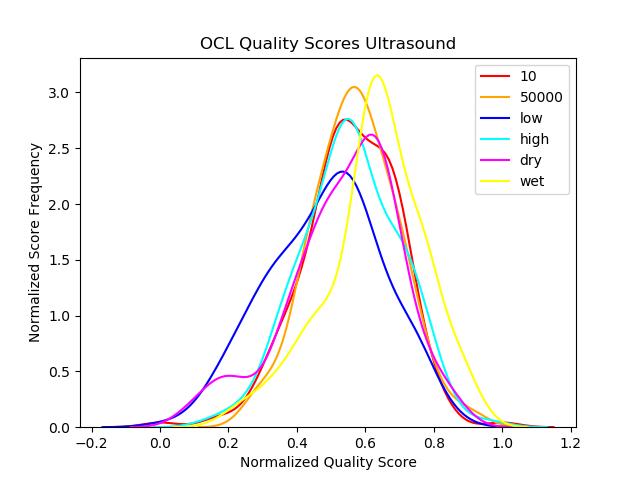}%
	\label{5case}}
	\hfil
	\subfloat[]{\includegraphics[width=0.33\textwidth]{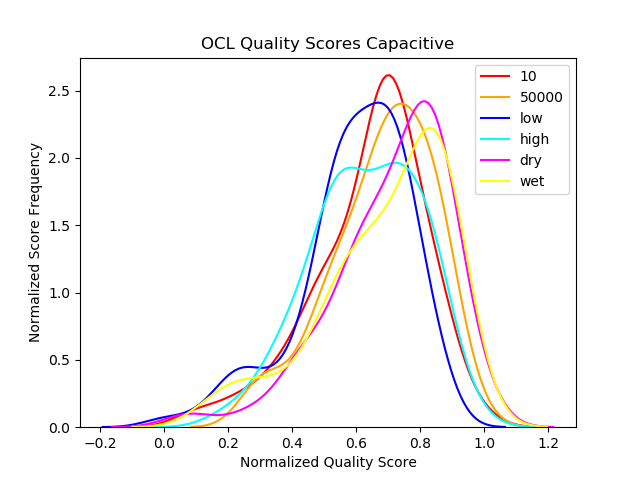}%
	\label{6case}}
	
	\subfloat[]{\includegraphics[width=0.33\textwidth]{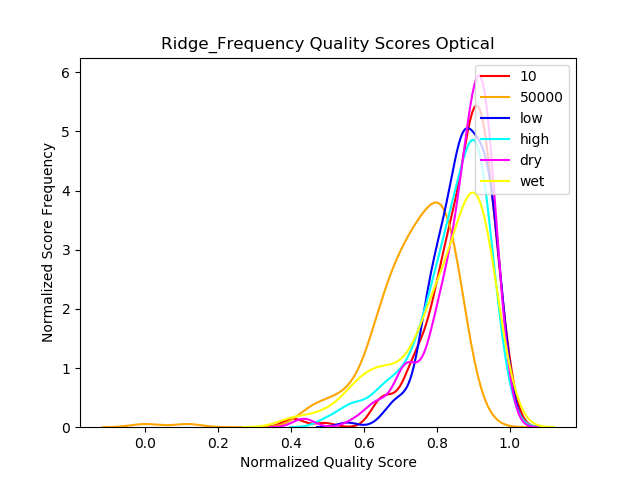}%
	\label{7case}}
	\hfil
	\subfloat[]{\includegraphics[width=0.33\textwidth]{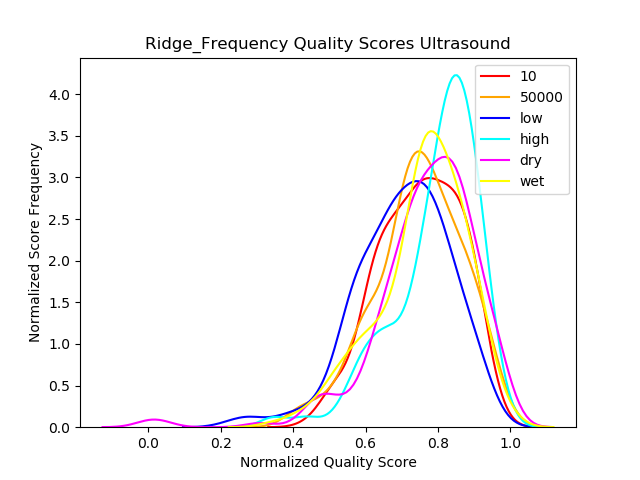}%
	\label{8case}}
	\hfil
	\subfloat[]{\includegraphics[width=0.33\textwidth]{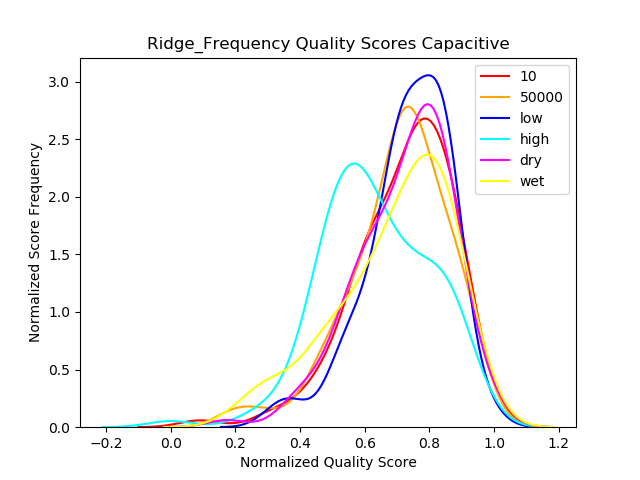}%
	\label{9case}}
	
	\hfil
	\subfloat[]{\includegraphics[width=0.33\textwidth]{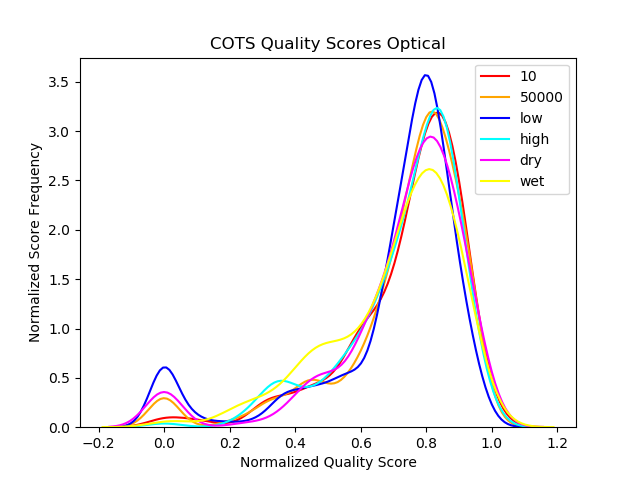}%
	\label{10case}}
	\hfil
	\subfloat[]{\includegraphics[width=0.33\textwidth]{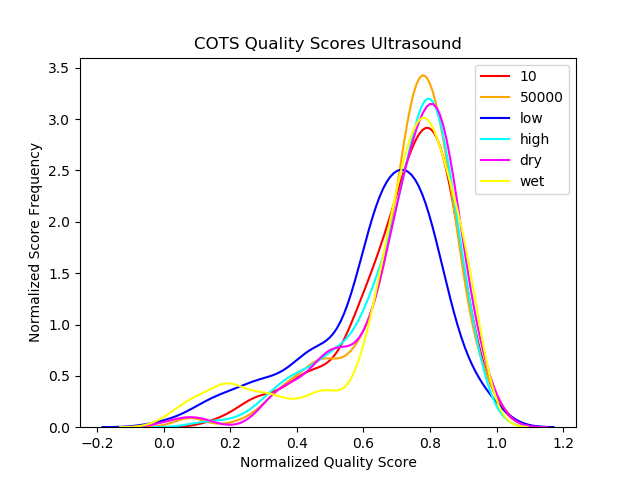}%
	\label{11case}}
	\hfil
	\subfloat[]{\includegraphics[width=0.33\textwidth]{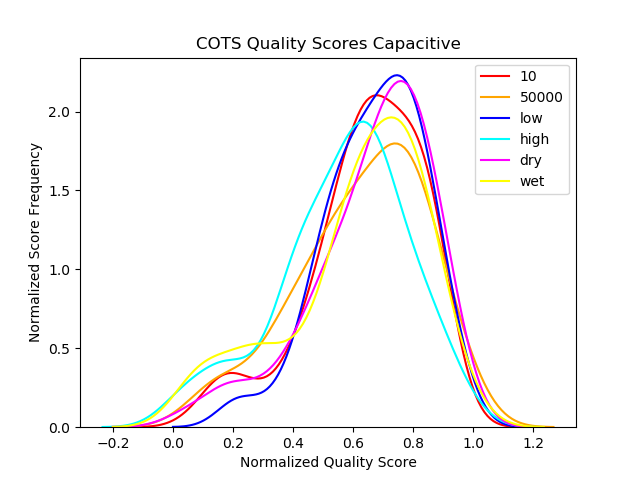}%
	\label{12case}}
	\caption{Quality-score distributions for each quality-metric on each of the three readers.}
	\label{fig:dist}
\end{figure*}

Once computed, the fingerprint quality-scores associated with each condition are subject to a statistical t-test at $\alpha = 0.05$ (since the quality-score distributions were approximately normally distributed) to determine which distributions of scores show statistically significant degradations in image quality captured for each device. Each condition-specific quality-score distribution is compared to the distribution of quality-scores computed from all fingerprint impressions in VCC (the complete quality-score distribution). We compare each condition-specific quality-score distribution to the complete quality-score distribution rather than to the quality-score distribution specific to normal capture impressions since the normal impressions were always captured first on each device (the lack of subject familiarity with fingerprint capture could bias the quality-score distribution stemming from the normal fingerprint impressions). The score distributions for each condition are shown in Figure~\ref{fig:dist}. These plots show the frequency of values (in the range $[0,1]$) for each quality-metric for impressions acquired under each of the different conditions, where values near 1 indicate higher quality. The results of the t-tests are shown in Table~\ref{tab:t_test} and show which perturbations result in statistically significant drops in the quality of the captured fingerprint images on each reader.

After computing all quality-score distributions and the subsequent t-tests, we conclude our white-box fingerprint evaluation with the aforementioned uncertainty analysis. In this evaluation, the feature sets, $S_k$, of the uncertainty analysis are the images of fingerprint impressions. The perturbed feature sets are then the impressions of the same finger captured under the varying conditions (e.g., pressure, illumination, and finger moisture). Finally, we measure the uncertainty in the scores $s_{k,n}$ obtained from each of the quality-metrics (Global Orientation Quality, Ridge Frequency, Orientation Certainty Level, and COTS). The uncertainty values for the fingerprint readers under each condition are given in Table~\ref{tab:uncertainty_readers}. 
        
% \begin{figure*}
% 	\centering
% 	\includegraphics[width=1\textwidth]{images/t_test.png}%
% 	\caption{t-test statistics for each quality-metric.}
% 	\label{fig:t_test}
% \end{figure*}

\begin{table*}[t]
\renewcommand{\arraystretch}{1.3}
\caption{t-test statistics for each quality-metric. Not highlighted: $t < 1.658$, highlighted in \textcolor{Yellow}{yellow}: $1.658 < t \leq 5$, highlighted in \textcolor{Orange}{orange}: $5 < t \geq 10$, highlighted in \textcolor{Red}{red}: $t > 10$.}
\label{tab:t_test}
\centering
\begin{tabular}{p{0.13\linewidth-2\tabcolsep}*{12}{C{0.07\linewidth-2\tabcolsep}}}
\toprule
 & \multicolumn{4}{c}{Optical} & \multicolumn{4}{c}{Ultrasound} & \multicolumn{4}{c}{Capacitive} \\
% \midrule
 & GOQ & RF & OCL & COTS & GOQ & RF & OCL & COTS & GOQ & RF & OCL & COTS \\
% \midrule
\cmidrule{1-1} \cmidrule(rl){2-5} \cmidrule(rl){6-9} \cmidrule(rl){10-13}
% \midrule
Dark lighting   & -2.83  & -4.28 & -0.54  & -1.48 & -2.33 & 0.35  & -0.25 & -0.81 & 0.15   & -1.23 & 0.97  & -0.64 \\
\cmidrule{1-1} \cmidrule(rl){2-5} \cmidrule(rl){6-9} \cmidrule(rl){10-13}
Bright lighting & -15.57 & \textcolor{Red}{10.30} & -16.63 & -0.43  & -2.72 & 0.59  & -1.44 & -1.55 & -2.22 & -0.17 & -1.43 & 0.14  \\
\cmidrule{1-1} \cmidrule(rl){2-5} \cmidrule(rl){6-9} \cmidrule(rl){10-13}
Low pressure    & \textcolor{Red}{10.71}  & -6.01  & \textcolor{Red}{14.30}  & \textcolor{Yellow}{1.68}   & \textcolor{Yellow}{1.85}  & \textcolor{Yellow}{4.88}  & \textcolor{Orange}{5.53}  & \textcolor{Yellow}{4.00}  & \textcolor{Orange}{5.08}  & -2.44  & \textcolor{Orange}{3.51}  & -2.38  \\
\cmidrule{1-1} \cmidrule(rl){2-5} \cmidrule(rl){6-9} \cmidrule(rl){10-13}
High pressure   & \textcolor{Yellow}{2.11}   & -1.27 & 0.35   & -1.21 & 1.29  & -4.34 & -0.16 & -1.04 & -3.59 & \textcolor{Yellow}{3.84}  & 1.47  & \textcolor{Yellow}{2.88}  \\
\cmidrule{1-1} \cmidrule(rl){2-5} \cmidrule(rl){6-9} \cmidrule(rl){10-13}
Dry finger      & \textcolor{Yellow}{2.45}   & -4.30 & \textcolor{Red}{10.99}  & -0.06  & 0.99  & -1.36 & 1.46  & -1.71 & -0.13 & -1.34 & -2.72 & -1.51 \\
\cmidrule{1-1} \cmidrule(rl){2-5} \cmidrule(rl){6-9} \cmidrule(rl){10-13}
Wet finger      & \textcolor{Yellow}{2.64}   & 1.49  & \textcolor{Yellow}{3.05}   & 1.01  & 0.05  & -0.35 & -6.07 & 0.34  & 0.47  & 0.58  & -1.59 & 0.81 \\
\bottomrule
\multicolumn{13}{l}{Note: $|t| > 1.658$ indicates a statistically significant difference between two population means for t-test with $120$ DOF at $95 \%$ confidence.}
\end{tabular}
% \vspace{-1.6em}
\end{table*}
        
\subsection{Observations}

For the optical-based reader, we see that there is a statistically significant drop in image quality associated with the conditions of varying pressure and finger moisture, as indicated by the Global Orientation Quality, Orientation Certainty Level, and COTS algorithms. Additionally, we see a statistically significant drop in quality for the Ridge Frequency (RF) metric on the impressions captured on the optical (FTIR) reader under bright lighting. We hypothesis that the reason the RF algorithm flags these impressions as poor quality, where the other orientation-based metrics did not, is due to the stronger dependence of the RF algorithm on the contrast between the dark and light pixels of the ridges and valleys, respectively. The orientation algorithms are not as sensitive to the absolute contrast between the ridges and valleys since they rely on the direction of the maximum principal component vectors within each block of the image and not the magnitude of these responses. 

% Therefore, since the ridges of the impressions captured under bright lighting appear to have a more median gray level value (e.g., pixel values around $100$), as opposed to the darker ridges (e.g., pixel values around $0$) seen in the impressions acquired under other conditions, the RF algorithm flags these impressions as low quality.
        
 For the ultrasound-based reader, we only see statistically significant drops in quality for the low pressure condition. Since this imaging technology is able to penetrate the surface of human skin and obtain sub-dermal images of the fingerprint ridge structure, they are expected to be robust to contaminants on the finger surface, (e.g., the moisture we applied to the fingers). However, due to a large impedance mismatch between air and human skin, these sensors are not robust to partial contact with the imaging surface, which is characteristic of impressions applied with low pressure.
        
 Finally, for the capacitive reader, there is a decrease in image quality seen in the low pressure and high pressure impressions. The orientation-based metrics flag the low pressure impressions as poor quality, whereas the RF and COTS's proprietary quality detection algorithm flag the high pressure impressions as poor. The orientation-based algorithms flag the low pressure impressions due to the inconsistent contact between the reader and the fingers of the subjects that are not being firmly pressed onto the imaging surface. Due to the elastic nature of human skin, the ridge lines of a fingerprint impression will appear thicker with increased pressure applied to the imaging platen, as detected by the RF and COTS algorithms. 
 
%  However, it is interesting that only in the case of the capacitive reader was this effect sufficient for any of the quality-metrics to flag these impressions as being of particularly low quality. It is possible that this discrepancy is due to the smaller acquisition window of the capacitive reader where only the central portion of the finger that is being firmly pressed onto the imaging surface is being captured, whereas the optical and ultrasound readers capture more of the outer edges of the fingerprint impression where the ridge width is less distorted due the force applied being concentrated toward the center of the finger.
 
 From the uncertainty analysis shown in Table~\ref{tab:uncertainty_readers}, we observe that the ultrasound reader, which was examined in this work, exhibited the lowest uncertainty in most of the quality-metrics evaluated, indicating that the quality of impressions captured by this reader demonstrate less variance due to adverse capture conditions of illumination, finger moisture, and contact pressure. This finding could be attributed to the robustness of ultrasound-based fingerprint sensors, as highlighted by our corresponding white-box evaluation experiments.

\section{Feature Extractor Evaluation}

\begin{figure}
	\centering
	\includegraphics[width=0.45\textwidth]{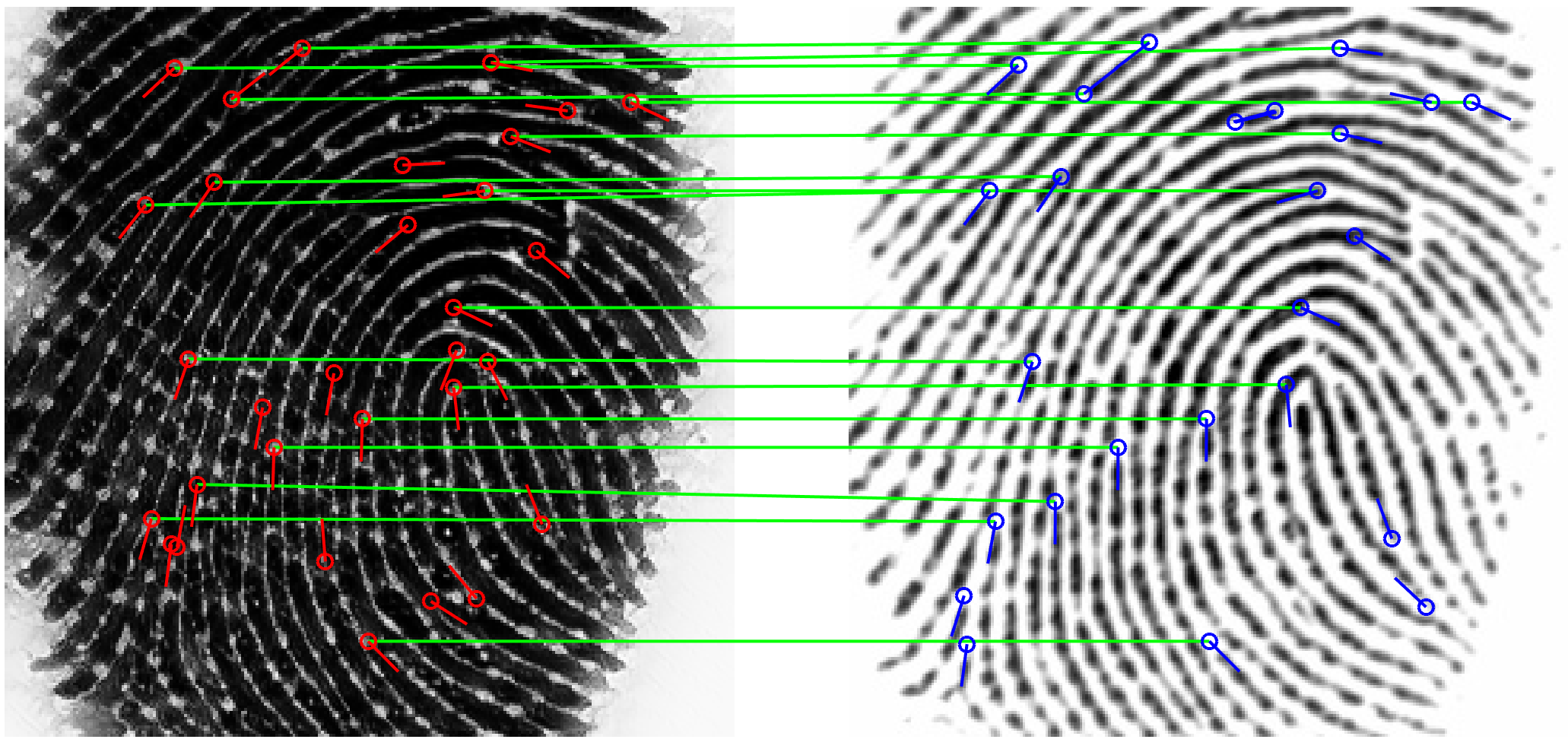}%
	\caption{Example of minutiae correspondence needed for the white-box feature extractor evaluation obtained using style transfer. The image on the right is an impression captured under normal capture conditions and the image on the left is the same impression after applying style transfer to reflect an impression captured under high pressure.}
	\label{fig:minu_corr}
\end{figure}

\begin{figure*}
	\centering
	\includegraphics[width=1\textwidth]{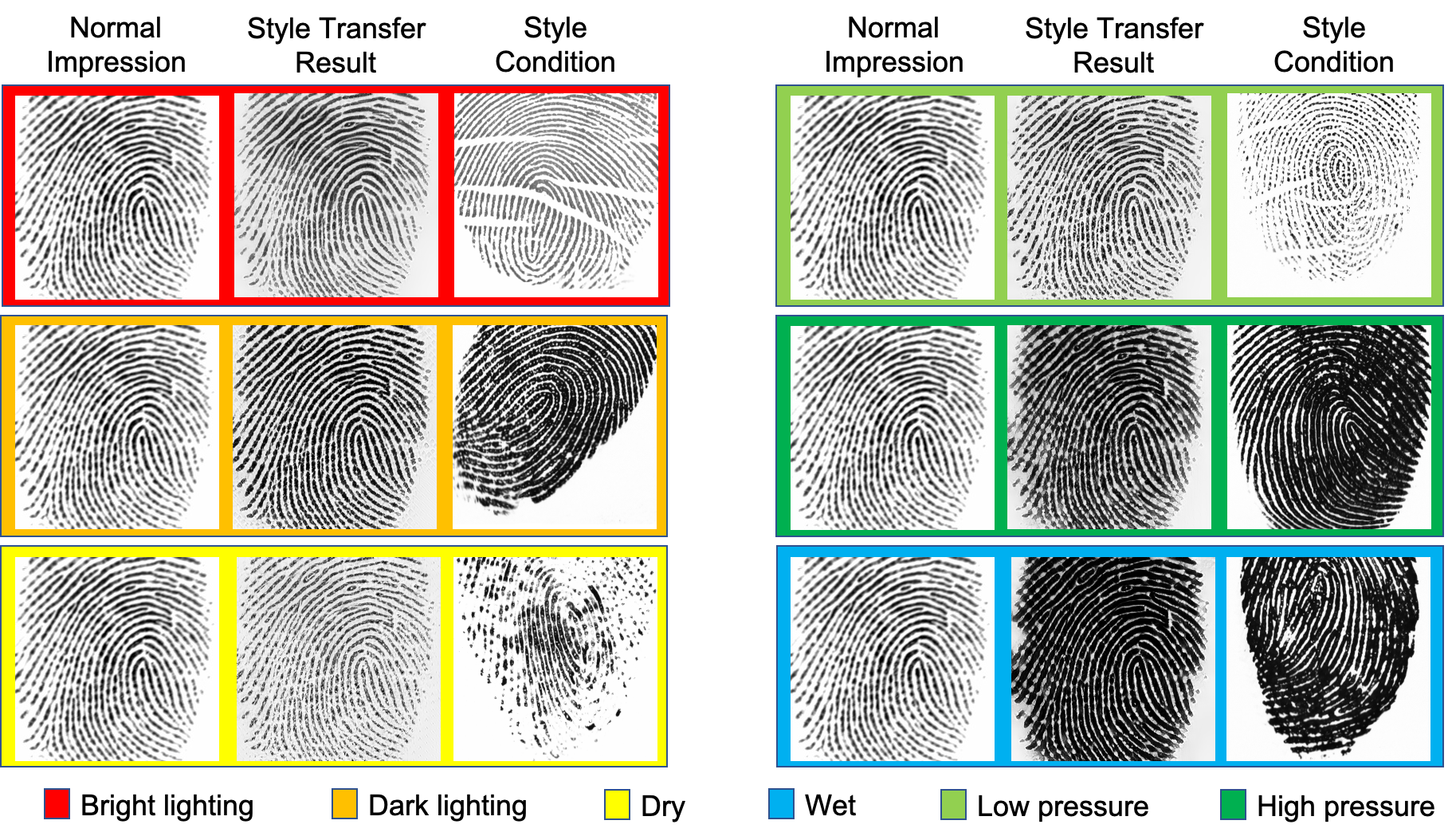}%
	\caption{Example stylized images.}
	\label{fig:stylized_imgs}
\end{figure*}

Our white-box evaluation of the minutiae extraction module follows a similar paradigm as that of~\cite{chugh2017benchmarking}. In particular, we first add perturbations to a fingerprint image for which we have manually annotated ground truth minutiae locations and orientations. Then, we pass the fingerprint through the minutiae extractor to see if the perturbation causes a degradation in the accuracy of the minutiae extractor. In~\cite{chugh2017benchmarking}, simple perturbation techniques such as motion blur and random noise were used as perturbations. Our white-box minutiae extractor evaluation is made more robust via the adoption of more realistic perturbations. 

To make realistic perturbations, we adopt the latest techniques of neural style transfer~\cite{huang2017arbitrary}, to transfer style images from our VCC dataset (\textit{e.g.} wet impression style) to ``content" fingerprint images for which we have ground truth minutiae locations (in our case, we use the manually annotated FVC 2002 DB1 A) (see Figure~\ref{fig:minu_corr}). Examples of style images from our VCC dataset, content images from FVC 2002 DB1 A, and style transferred images (which we use to evaluate the minutiae extractors) are shown in Figure~\ref{fig:stylized_imgs}.
        
We train the fingerprint style transfer network proposed in~\cite{chugh2020fingerprint} and extended in~\cite{grosz2020fingerprint} with the aggregated dataset of $16\,731$ optical (FTIR) captured images. Once the network is trained and sufficiently adept at transferring styles between fingerprint images, we apply the network to produce our synthetic database of stylized images. For content images (or normal impressions) we select the top 100 normal impressions from FVC 2002 DB1 A that have the highest NFIQ 2.0 scores. This ensures that the content (or normal) impressions used have the highest fidelity, so that performance degradation of the feature extractors can be mostly attributed to the specific stylized conditions. Next, we manually select, from VCC, the 20 most representative impressions of each adverse capture condition as our style images. Finally, we use these content and style images to generate 1000 stylized images for each perturbation condition resulting in a total of 6000 stylized, fingerprint images. Figure~\ref{fig:stylized_imgs} shows the result of the style transfer using exemplar images of each subset of conditions. 
        
\begin{table*}[]
\centering
\caption{Detection and Localization Statistics of the Two COTS Minutiae Feature Extractors for the Different Capture Conditions on the FTIR Optical-Based Reader. Bold values indicate worst performance.}
\label{tab:minu_stats}
\footnotesize
{
\begin{tabular}{p{0.135\textwidth}p{0.06\textwidth}*{6}{p{0.092\textwidth}}p{0.035\textwidth}}
 & Minutiae \newline Extractor & Bright \newline Lighting & Dark \newline Lighting  & Low \newline Pressure   & High \newline Pressure  & Wet \newline Finger   & Dry \newline Finger & \textbf{Avg.}\\
\midrule
\multirow{2}{0.135\textwidth}{Paired Minutiae ($P_i / M_i$) Avg. (s.d.)} & COTS-A & 0.576 (0.086) & 0.574 (0.091) & 0.584 (0.086) & 0.575 (0.091) & \textbf{0.573 (0.088)} & 0.577 (0.087) & 0.58\\
 & COTS-B & \textbf{0.607 (0.088)} & 0.613 (0.089) & 0.612 (0.077) & 0.609 (0.088) & 0.609 (0.083) & 0.615 (0.091) & 0.61\\
\midrule 
\multirow{2}{0.135\textwidth}{Missing Minutiae ($I_i / M_i$) Avg. (s.d.)} & COTS-A & 0.424 (0.086) & 0.426 (0.091) & 0.416 (0.086) & 0.426 (0.091) & \textbf{0.427 (0.088)} & 0.423 (0.087) & 0.42\\
 & COTS-B & \textbf{0.393 (0.088)} & 0.387 (0.089) & 0.379 (0.077) & 0.391 (0.088) & 0.391 (0.083) & 0.386 (0.091) & 0.39\\
\midrule
\multirow{2}{0.135\textwidth}{Spurious Minutiae ($D_i / M_i$) Avg. (s.d.)} & COTS-A & 0.189 (0.071) & 0.191 (0.071) & \textbf{0.198 (0.068)} & 0.190 (0.073) & 0.191 (0.073) & 0.193 (0.071) & 0.19\\
 & COTS-B & 0.207 (0.065) & 0.205 (0.068) & 0.213 (0.063) & 0.208 (0.068) & \textbf{0.215 (0.067)} & 0.208 (0.066) & 0.21\\
\midrule
\multirow{2}{0.12\textwidth}{Goodness Index Avg. (s.d.)} & COTS-A & -0.033 (0.174) & -0.037 (0.185) & -0.026 (0.176) & -0.037 (0.188) & \textbf{-0.040 (0.180)} & -0.035 (0.178) & -0.03\\
 & COTS-B & 0.006 (0.182) & 0.019 (0.190) & 0.026 (0.168) & 0.010 (0.184) & \textbf{0.003 (0.177)} & 0.018 (0.191) & 0.01\\
\midrule
\multirow{2}{0.14\textwidth}{Positional Error ($e_p$) (pixels) Avg. (s.d.)} & COTS-A & 3.63 (0.790) & 3.50 (0.809) & 3.67 (0.802) & 3.56 (0.805) & 3.57 (0.799) & \textbf{3.69 (0.856)} & 3.60\\
 & COTS-B & 3.62 (0.699) & 3.60 (0.719) & 3.66 (0.725) & 3.62 (0.726) & 3.65 (0.747) & \textbf{3.67 (0.732)} & 3.64\\
\midrule
\multirow{2}{0.15\textwidth}{Orientation Error ($e_\theta$) (rad) Avg. (s.d.)} & COTS-A & 0.197 (0.223) & 0.178 (0.207) & \textbf{0.237 (0.251)} & 0.209 (0.234) & 0.218 (0.247) & 0.228 (0.257) & 0.21\\
 & COTS-B & 0.220 (0.247) & 0.211 (0.237) & 0.221 (0.246) & 0.226 (0.247) & \textbf{0.243 (0.259)} & 0.234 (0.253) & 0.23\\
\bottomrule
\end{tabular}
}
\end{table*}
    
\subsection{Evaluation Metrics}
 
After using style transfer to apply realistic perturbations from VCC to the FVC 2002 DB1 A dataset (for which we have manually marked minutiae), we evaluate two COTS minutiae extractors by computing the number of missing and spurious minutiae (Goodness Index) and the error in $x$, $y$, and $\theta$ (Positional Error) from the ground truth annotations. We conclude the evaluation with an uncertainty analysis of the minutiae extractors.
        
\subsubsection{Goodness Index}
Given a fingerprint image, let $F_g = \{f_g^1,f_g^2,...,f_g^M\}$ be the set of $M$ manually marked ground truth minutiae and $F_d = \{f_d^1,f_d^2,...,f_d^N\}$ be the set of $N$ minutiae detected by a given minutiae extractor. Evaluating the detection performance of the minutiae feature extractors requires establishing a correspondence between minutiae points detected in set $F_g$ and set $F_d$, in which a minutiae in one set is said to be paired with a minutiae in the other set if the distance between the two minutiae locations lies within a distance threshold $\delta$. Empirically, the average ridge width of a $500$ dpi fingerprint impressions is found to be $9$ pixels~\cite{maltoni2009handbook}; therefore, an appropriate choice for $\delta$ is $10$ pixels. If multiple pairs fall within the threshold, the pair with the closest distance to the ground truth annotation is chosen, where ties are broken in favor of the pair with the smallest orientation difference. Finally, a score is assigned to assess the detection performance following the Goodness Index (GI) introduced by Ratha \textit{et al.} in \cite{ratha1995adaptive}: 
            
            \begin{equation}
                GI = \frac{\sum_{i=1}^L[P_i - D_i - I_i]}{\sum_{i=1}^L M_i}
            \end{equation}
            
            where $L =$ number of $16\times16$ non-overlapping patches in the input image, $P_i =$ number of paired minutiae in the $i^{th}$ patch, $D_i =$ number of spurious minutiae in the $i^{th}$ patch, $D_i \leq 2 \cdot M_i$, $I_i =$ number of missing minutiae in the $i^{th}$ patch, and $M_i =$ number of ground truth minutiae in the $i^{th}$ patch, $M_i > 0$. To mitigate the effect of outlier patches, the number of spurious minutiae ($D_i$) in a patch is restricted to a maximum value of $2 \cdot M_i$. The range of the Goodness Index is $[-3,1]$, where larger values indicate better minutiae extractor performance.
            
\subsubsection{Positional Error}
           
            The positional error ($e_p$) between a set of $P$ detected minutiae, $\hat{f}_d = \{\hat{f}^1_d,\hat{f}^2_d,...,\hat{f}^P_d\}$ and a paired subset of ground truth minutiae, $\hat{f}_g = \{\hat{f}^1_g,\hat{f}^2_g,...,\hat{f}^P_g\}$, with $\hat{f}_d \in F_d$ and $\hat{f}_g \in F_g$, is computed via the Root Mean Squared Deviation (RMSD) \cite{turroni2011improving}:
            
            \begin{equation}
                e_p(\hat{f}_d,\hat{f}_g) = \sqrt{\frac{\sum_{i=1}^P[(x_g^i - x_d^i)^2 + (y_g^i - y_d^i)^2]}{P}}
            \end{equation}
            
            where $(x_d^i,y_d^i)$ and $(x_g^i,y_g^i)$ represent the locations of the detected and ground truth minutiae, respectively. Additionally, the orientation error ($e_\theta$) between a set of paired minutiae is given by:
            
            \begin{equation}
                e_\theta(\hat{f}_d,\hat{f}_g) = \sqrt{\frac{\sum_{i=1}^P\Phi(\theta_g^i,\theta_d^i)^2}{P}}
            \end{equation}
            
            where, 
            
            \begin{equation}
                \Phi(\theta_1,\theta_2) = 
                \begin{dcases}
                    \theta_1 - \theta_2, & -\pi \leq \theta_1 - \theta_2 < \pi \\
                    2\pi + \theta_1 - \theta_2, & \theta_1 - \theta_2 < -\pi \\
                    -2\pi + \theta_1 - \theta_2, & -\pi \leq \theta_1 - \theta_2 \geq \pi 
                \end{dcases}
            \end{equation}

 \subsection{Uncertainty Analysis}           
We measure the uncertainty of each COTS minutiae feature extractor from the Goodness Index scores between minutiae detected by each feature extractor and the ground truth minutiae of each normal impression. Thus, the feature sets, $S_k$ for this evaluation are the ground truth minutiae locations of the normal capture impressions, the perturbed feature sets, $S'_{k,n}$, are the minutiae feature sets output by each COTS feature extractor on the style transferred impressions of varying moisture, illumination, and pressure, and the evaluation scores, $s_{k,n}$, are the values given by the goodness index. The resulting uncertainty values are given in Table~\ref{tab:uncertainty_feat}.

\begin{table}
\renewcommand{\arraystretch}{1.3}
\caption{Uncertainty Scores for COTS-A and COTS-B Minutiae Feature Extractors.}
\label{tab:uncertainty_feat}
\centering
\begin{tabular}{m{0.4\linewidth-2\tabcolsep}>{\centering\arraybackslash}m{0.3\linewidth-2\tabcolsep}>{\centering\arraybackslash}m{0.3\linewidth-2\tabcolsep}}
\toprule
 & COTS-A & COTS-B \\
\midrule
Finger Moisture & 0.0153 & 0.0126 \\
\midrule
Contact Pressure & 0.0105 & 0.0137 \\
\midrule
Illumination & 0.0126 & 0.0146 \\
\bottomrule
\end{tabular}
% \vspace{-1.6em}
\end{table}

\subsection{Observations}
The results of the detection and localization experiments are shown in Table~\ref{tab:minu_stats}. Fingerprint impressions captured with wet skin appear to score the worst in terms of Goodness Index, presumably due to the high ratio of missing minutiae and corresponding low number of paired minutiae compared to the unperturbed ground truth minutiae locations. Dry fingers and low pressure impressions also stand out as problematic for both COTS feature extractors for their high values of localization errors, \textit{i.e.}, positional and orientation errors of the detected minutiae points. These discrepancies are likely due to the high ratio of spurious minutiae that are characteristic of these capture conditions caused by the inconsistent contact between the finger and the imaging surface. This inconsistent contact leads to artificial breaks in the fingerprint ridge structure that results in many spurious minutiae. Finally, we observe that COTS-B outputs more paired minutiae compared with the ground truth templates on average; however, in doing so produces more spurious minutiae than COTS-A. So, even though COTS-B achieves a higher average Goodness Index than COTS-A, it suffers from higher localization errors in terms of position and orientation of the detected minutiae. These findings could prove useful to an end user in selecting a feature extractor tailored to the security requirements of their application. 

% For example, high risk applications may favor COTS-A, which produces less spurious minutiae than COTS-B, to lower the chance of granting access to imposters (since many spurious minutiae may lead to increased imposter match scores). 
        
From the uncertainty evaluation results shown in Table~\ref{tab:uncertainty_feat} we note that COTS-A has lower uncertainty for impressions captured under varying contact pressure and illumination, whereas COTS-B has lower uncertainty on impressions obtained from varying finger moisture.

\begin{figure}
	\centering
	\subfloat[]{\includegraphics[width=0.45\textwidth]{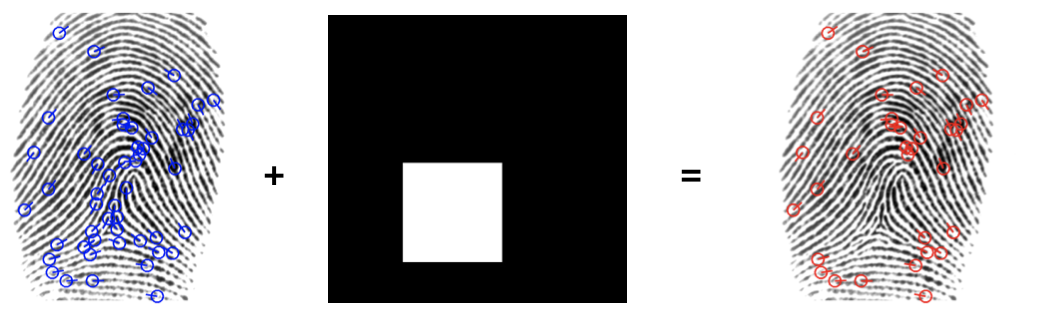}%
	\label{occlusion}}
	\hfil
	\subfloat[]{\includegraphics[width=0.45\textwidth]{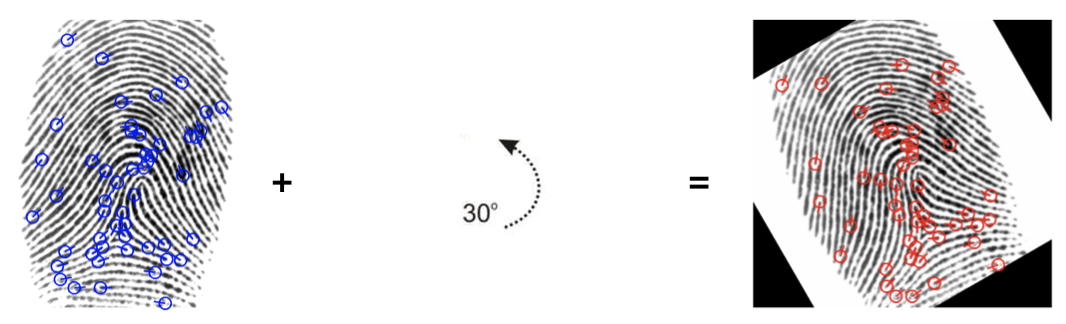}%
	\label{rotations}}
	\caption{Illustration of perturbation techniques to the input minutiae feature sets: a.) occlusion of minutiae via spatially contiguous blocks and b.) global rotation of minutiae locations.}
	\label{fig:perturbations}
\end{figure}

\begin{figure*}
	\centering
	\subfloat[]{\includegraphics[width=0.45\textwidth]{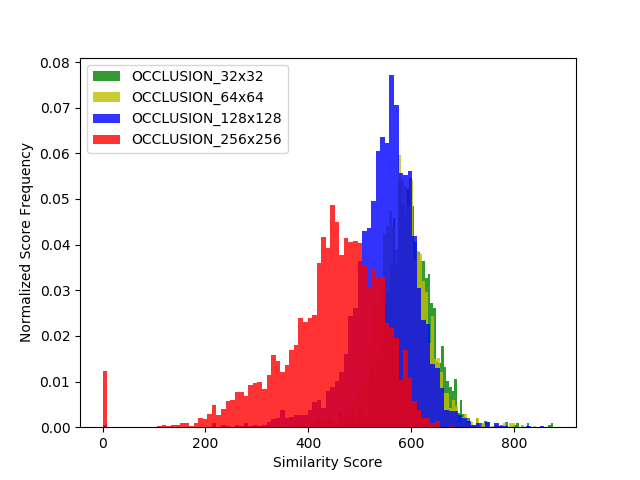}%
	\label{innovatrics_occlusion}}
% 	\hfil
	\subfloat[]{\includegraphics[width=0.45\textwidth]{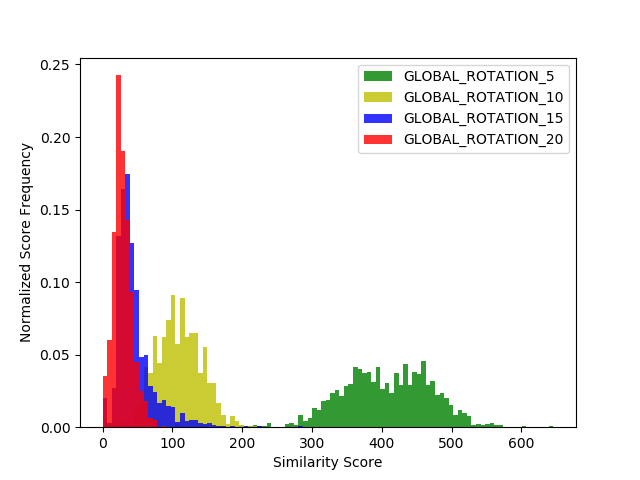}%
	\label{innovatrics_rotation}}
	\hfil
	\vspace{-1.0em}
	\subfloat[]{\includegraphics[width=0.45\textwidth]{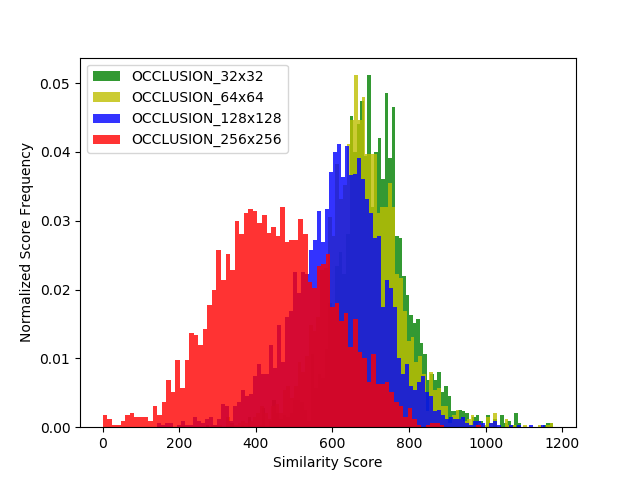}%
	\label{verifinger_occlusion}}
% 	\hfil
	\subfloat[]{\includegraphics[width=0.45\textwidth]{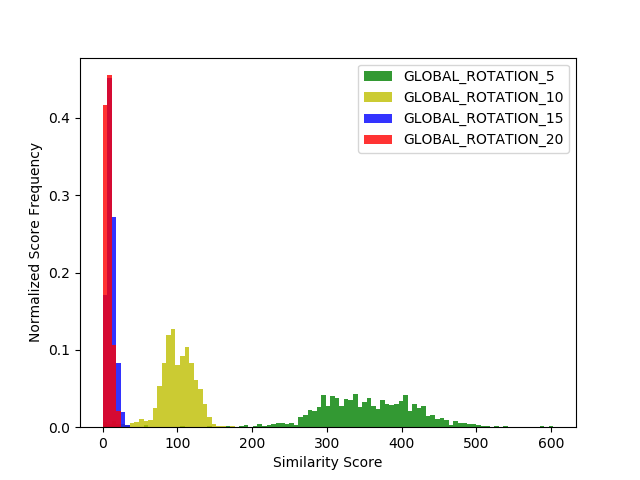}%
	\label{verifinger_rotation}}
	\caption{Genuine match score distributions for COTS-A and COTS-B for genuine fingerprint impression pairs subject to varying degrees of rotation and occlusion. (a) COTS-A subject to occlusion, (b) COTS-A subject to global rotation, (c) COTS-B subject to occlusion, and (d) COTS-B subject to global rotation.}
	\label{fig:matcher_eval}
	\vspace{-1.0em}
\end{figure*}

% \begin{figure*}
% 	\centering
% 	\subfloat[]{\includegraphics[width=0.45\textwidth]{images/matcher_eval/inn_fnmr_occlusion.png}%
% 	\label{innovatrics_occlusion_FNMR}}
% % 	\hfil
% 	\subfloat[]{\includegraphics[width=0.45\textwidth]{images/matcher_eval/inn_fnmr_rotation.png}%
% 	\label{innovatrics_rotation_FNMR}}
% 	\hfil
% 	\vspace{-1.0em}
% 	\subfloat[]{\includegraphics[width=0.45\textwidth]{images/matcher_eval/veri_fnmr_occlusion.png}%
% 	\label{verifinger_occlusion_FNMR}}
% % 	\hfil
% 	\subfloat[]{\includegraphics[width=0.45\textwidth]{images/matcher_eval/veri_fnmr_rotation.png}%
% 	\label{verifinger_rotation_FNMR}}
% 	\caption{FNMR vs. increasing perturbations of rotation and occlusion of input minutiae feature sets for two COTS fingerprint matching algorithms. (a) COTS-A subject to occlusion, (b) COTS-A subject to global rotation, (c) COTS-B subject to occlusion, and (d) COTS-B subject to global rotation.}
% 	\label{fig:matcher_eval_FNMR}
% 	\vspace{-1.0em}
% \end{figure*}

\begin{table}
\renewcommand{\arraystretch}{1.3}
\caption{FNMR (\%) at FAR = $0.1 \%$ vs. increasing perturbations of rotation and occlusion of input minutiae feature sets for two COTS fingerprint matching algorithms.}
\label{tab:matcher_eval_FNMR}
\centering
\begin{tabular}{m{0.15\linewidth-2\tabcolsep}m{0.17\linewidth-2\tabcolsep}*{2}{>{\centering\arraybackslash}m{0.15\linewidth-2\tabcolsep}}*{2}{>{\centering\arraybackslash}m{0.19\linewidth-2\tabcolsep}}}
\toprule
($pixels^2$) & & $32\times32$ & $64\times64$ & $128\times128$ & $256\times256$ \\
\midrule
\multirow{2}{0.15\textwidth}{Occlusion} & COTS-A & 0.00 & 0.00 & 0.06 & 1.23 \\
 & COTS-B & 0.00 & 0.00 & 0.00 & 0.31 \\
\midrule
($degrees$) & & $5$ & $10$ & $15$ & $20$ \\
\midrule
\multirow{2}{0.15\textwidth}{Rotation} & COTS-A & 0.29 & 1.41 & 25.34 & 58.74 \\
 & COTS-B & 0.23 & 3.06 & 100.00 & 99.94 \\
\bottomrule
\end{tabular}
% \vspace{-1.6em}
\end{table}

\section{Matcher Evaluation}
    
 To evaluate minutiae matchers, we again utilize the manually annotated FVC 2002 DB1 A database. In particular, we perturb the ground truth FVC minutiae sets to generate a large database of perturbed minutiae sets. Then, we compare the match scores computed between the unperturbed minutiae sets and the corresponding perturbed minutiae sets via their score distributions and also an uncertainty analysis. This enables us to determine the sensitivity of the matcher modules to our perturbation techniques.
    
The perturbations we explore extend those of our previous study in~\cite{grosz2019whitebox} (random $x$, $y$, and $\theta$ displacements of minutiae locations, random removal/addition of minutiae, and non-linear distortion of minutiae locations) by: (i) removing minutiae within randomly occluded blocks and (ii) global rotation of all minutiae points. These perturbations were chosen to better model the types of minutiae perturbations which the minutiae extractors will be exposed to in an operational setting. In wet fingerprints, the ridge structure of the fingerprint collapses in spatially contiguous blocks (resulting in blocks of missing minutiae). As subjects place their finger on the fingerprint reader, different presentation angles will cause the minutiae points to be a different global orientations.

%Our procedure to compare the genuine similarity scores between corresponding unperturbed and perturbed impressions is as follows: 
%        
%        \begin{enumerate}
%            \item Obtain $M = 800$ $A_{ref,k}$ reference fingerprint impressions from the publicly available FVC 2002 DB1A fingerprint database~\cite{maio2002fvc2002}.
%            \item Obtain $M$ minutiae feature sets, $S_{ref,k}$, from each $A_{ref,k}$.
%            \item For each $S_{ref,k}$, synthesize $N = 5$ perturbed minutiae sets, $S'_{test,k,n}$, $1 \leq n \leq N$. %In particular, $S'_{test,k,n} = S_{ref,k} + \Delta_{test,n}$.
%            \item Generate genuine similarity scores, $s_{k,n}$, between $S_{ref,k}$ and each $S'_{test,k,n}$ using two COTS minutiae-based matchers.
%            \item Normalize the scores, $s_{k,n}$, to be in the range of $[0,1]$ using min-max normalization, where the min and max are matcher specific values.
%            \item Repeat steps $1$ to $7$ for each perturbation type and perturbation parameters.
%        \end{enumerate}
        
\subsection{Observations}

Genuine similarity score distributions for both COTS matchers on fingerprint impressions subject to increasing levels of minutiae perturbations of global rotation and occlusion are shown in Figure~\ref{fig:matcher_eval}. Specifically, we evaluate the robustness of each COTS matcher to rotations of ($5$, $10$, $15$, and $20$) degrees clockwise and counter-clockwise and occlusions due to random sized boxes of increasing area from ($32\times32$, $64\times64$, $128\times128$, and $256\times256$) pixels.
        
From Figure~\ref{fig:matcher_eval}, we observe that global rotation greater than $10$ degrees is detrimental to genuine similarity scores of both matchers. Indeed, it is possible that the feature extraction module of each COTS matcher (or other minutiae-based fingerprint recognition systems) performs an alignment step to mitigate variation in presentation angle of unconstrained capture scenarios; however, it is clear from this experiment that both of these COTS matchers are not robust to possible alignment errors. Additionally, we observe degradations to genuine similarity scores of both matchers due to increasing occlusion area; albeit, not as significant of a performance decline due to global rotation.
    
Of course, slight drops in genuine similarity scores output by the matcher module are not expected to significantly degrade recognition unless the scores fall below a certain threshold, at which they are classified as a non-match. This threshold is either set by the system designer or by end-users to satisfy security constraints of a particular application. In our evaluation, we select the thresholds recommended by the manufacturers of both COTS systems at a False Acceptance Rate (FAR) of $0.1 \%$ and compute the false non-match rate (FNMR) due to the increasing perturbations. The FNMR vs. increasing rotation and occlusion are shown in Table~\ref{tab:matcher_eval_FNMR}. From these results, we note that global rotation quickly leads to poor performance of both COTS matchers, whereas missing minutiae due to occluded blocks in the fingerprint impressions leads to only a slight degradation in FNMR of both systems. Interestingly, COTS-A demonstrates greater robustness to global rotation, but slightly worse robustness to occlusion compared to COTS-B.
 
Finally, we compute the uncertainty on the similarity scores due to the perturbations of global rotation and occlusion on the minutiae feature sets input to the matchers. In this case, the reference feature sets, $S_k$, are the unperturbed minutiae, the minutiae after applying the global rotation or occlusion are the perturbed feature sets, $S'_{k,n}$, and the matcher similarity scores between the unperturbed and perturbed feature sets are the evaluation scores, $s_{k,n}$. The uncertainty values for each COTS matcher are given in Table~\ref{tab:uncertainty_matchers}. We observe that COTS-A has lower uncertainty in the match scores produced on impressions perturbed with global rotation and comparable uncertainty to COTS-B on occluded impressions.

\begin{table}
\renewcommand{\arraystretch}{1.3}
\caption{Uncertainty Scores for COTS-A and COTS-B Matchers.}
\label{tab:uncertainty_matchers}
\centering
\begin{tabular}{m{0.4\linewidth-2\tabcolsep}>{\centering\arraybackslash}m{0.3\linewidth-2\tabcolsep}>{\centering\arraybackslash}m{0.3\linewidth-2\tabcolsep}}
\toprule
 & COTS-A & COTS-B \\
\midrule
Global Rotation & 0.0584 & 0.0576 \\
\midrule
Occlusion & 0.0077 & 0.0115 \\
\bottomrule
\end{tabular}
\end{table}

\section{Full-System Black-Box Evaluation}

In this section, we describe a black-box evaluation by computing match scores on fingerprint impressions that are first captured by the fingerprint reader, processed by a minutiae feature extractor, and finally passed to the matcher. For this evaluation we use fingerprint images from VCC captured on the FTIR optical reader under varying pressure, humidity, and lighting. We augment this dataset with images from FVC 2004 DB1A by labeling images from FVC 2004 DB1 A into one of our VCC adverse conditions via a classifier (\textit{i.e.} a condition classifier) which has been trained to do so on the VCC dataset. When computing match scores, genuine pairs are formed between normal captured impressions and the corresponding impression under an adverse condition.

\subsection{Observations}
Table~\ref{tab:black_box_fnmr} gives the FNMR of each COTS system (feature extractor and matcher) on impressions obtained under each of the varying capture conditions. From this table, we observe that varying lighting capture conditions for the optical-based reader cause very little, if any, degradation to matching performance for both COTS fingerprint systems. In contrast, impressions captured with wet fingers and high pressure significantly degrade matching performance, with wet impressions leading to the lowest scores. Additionally, high and low pressure impressions slightly increase the FNMR of both systems. Overall, COTS-B demonstrates slightly better robustness for fingerprint impressions captured under the adverse capture conditions included in this study.
        
These results seem to agree with the findings of the previous white-box evaluation of minutiae-based matchers, which showed that non-linear distortion and missing minutiae may greatly degrade the match scores of minutiae-based matchers~\cite{grosz2019whitebox}. This is evident in that the significant degradation due to wet impressions is likely due to the exaggerated number of minutiae occluded by large blobs of moisture on the finger surface. Furthermore, the increased pressure with which subjects present their fingers to the imaging surface in the high pressure impressions leads to severe distortion of the elastic fingerprint ridge structure. Lastly, that dry and low pressure impressions yield similar results is not surprising given the similar characteristics of these impressions which make them hard to distinguish visually.

\begin{table}
	\centering
	\caption{FNMR (\%) at FAR = $0.1 \%$ for Both COTS Matchers on the Various Capture Conditions.}
	\label{tab:black_box_fnmr}
	\small 
	{
	\begin{tabular}{p{0.1\textwidth}C{0.15\textwidth}C{0.15\textwidth}}
		\toprule
      		& COTS-A & COTS-B \\
		\midrule
	    Dry finger & 9.52 & 8.11 \\
		\midrule
	    Wet finger & 12.1 & 9.89 \\
		\midrule
	    Low pressure & 1.08 & 1.18 \\
		\midrule
	    High pressure & 1.70 & 0.92 \\
		\midrule
	    Bright lighting & 0.00 & 0.00 \\
		\midrule
	    Dark lighting & 0.00 & 0.00 \\
		\bottomrule
	\end{tabular}
	}
\end{table}

\section{Conclusion and Future Work}

In this work, we've proposed a framework for assessing performance of automated fingerprint recognition systems in both a black-box and white-box manner, thereby combining the strengths of each evaluation. Black-box evaluations provide a concise measure of recognition accuracy (i.e., overall system performance) to help end-users quickly determine if a system meets their application requirements; however, black-box evaluations do not give insight into the performance of each sub-module of the system suitable for comparing the various internal algorithms at each stage in the pipeline (i.e., image acquisition, feature extraction, and matching). This framework extends previous white-box evaluations of fingerprint readers, feature extractors, and matching algorithms which researchers and engineers can use to design better systems and compare algorithms. Our neural style transfer approach to augment existing fingerprint databases should also promote controlled robustness studies and aide the development of better algorithms. 

\section*{Acknowledgment}
The authors would like to thank the subjects in our study who are graduate researchers in the following laboratories at MSU with whom we have active collaboration: Computer Vision lab, Integrated Pattern Recognition and Biometrics lab, and Human Analysis Lab. We would also like to thank our collaborating researcher Nicholas G. Paulter for his valuable insights in suggesting this line of research.

\ifCLASSOPTIONcaptionsoff
  \newpage
\fi




\bibliographystyle{IEEEtran}
\bibliography{citations.bib}

\begin{IEEEbiography}[{\includegraphics[width=1in,height=1.25in,clip,keepaspectratio]{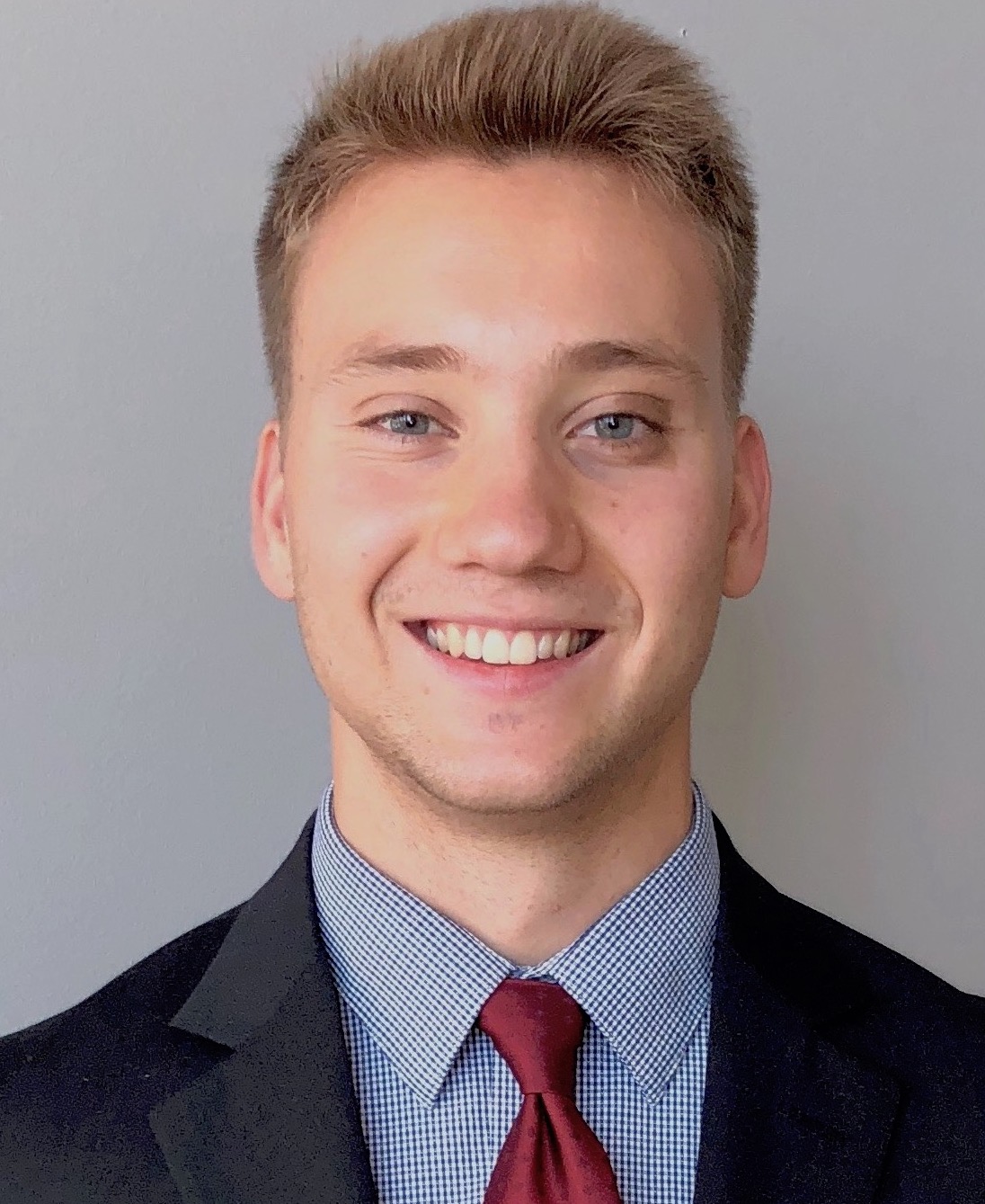}}]{Steven A. Grosz}
received his B.S. degree in Electrical Engineering from Michigan State University, East Lansing, Michigan, in 2019. He is currently a doctoral student in the Department of Computer Science and Engineering at Michigan State University. His primary research interests are in the areas of machine learning and computer vision with applications in biometrics.
\end{IEEEbiography}

\begin{IEEEbiography}[{\includegraphics[width=1in,height=1.25in,clip,keepaspectratio]{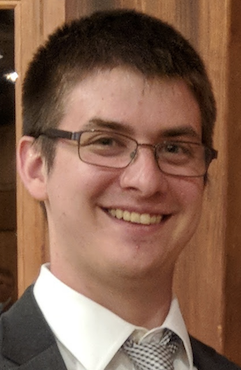}}]{Joshua J. Engelsma}
graduated magna cum
laude with a B.S. degree in computer science from Grand Valley State University, Allendale, Michigan, in 2016. He is currently working towards a PhD degree in the Department of Computer Science and Engineering at Michigan State University, East Lansing, Michigan. His research interests include pattern recognition, computer vision, and image processing with applications in biometrics. He won best paper award at ICB 2019 and is a student member of IEEE.
\end{IEEEbiography}

% \begin{IEEEbiography}[{\includegraphics[width=1in,height=1.25in,clip,keepaspectratio]{img/nick_paulter}}]{Nicholas G. Paulter Jr.}
% is the Group Leader for the Security Technologies Group at NIST in Gaithersburg, MD. He develops and oversees metrology programs related to concealed weapon and contraband imaging and detection, biometrics for identification, and body armor characterization. He has authored or co-authored over 100 peer-reviewed technical articles and provided numerous presentations at a variety of technical conferences. He is a 2008-2009 Commerce Science and Technology Fellow and a 2010 IEEE Fellow
% \end{IEEEbiography}

\begin{IEEEbiography}[{\includegraphics[width=1in,height=1.25in,clip,keepaspectratio]{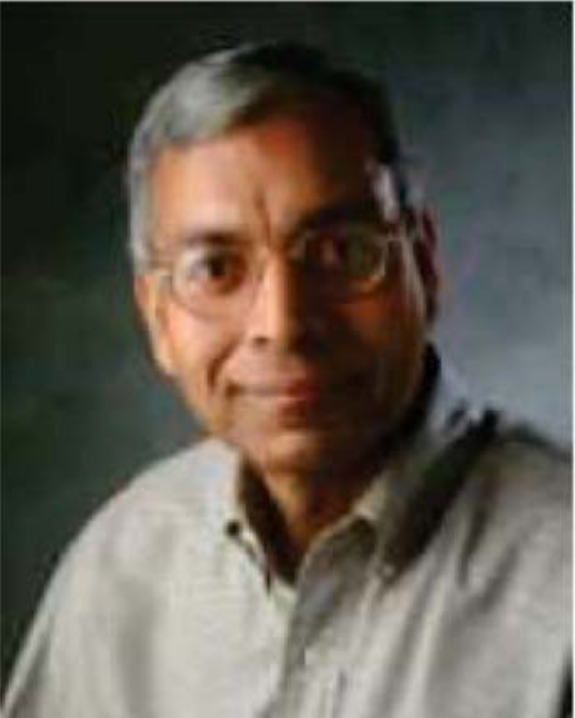}}]{Anil K. Jain}
is a University distinguished professor in the Department of Computer Science and Engineering at Michigan State University. His research interests include pattern recognition and biometric authentication. He served as the editor-in-chief of the IEEE Transactions on Pattern Analysis and Machine Intelligence. He is a member of the United States National Academy of Engineering and a Foreign Fellow of the Indian National Academy of Engineering.
\end{IEEEbiography}

\end{document}